\documentclass[sigconf,nonacm]{acmart}

\AtBeginDocument{%
  }

\setcopyright{none}
\renewcommand\footnotetextcopyrightpermission[1]{}

\usepackage{algorithm}
\usepackage{algorithmic}
\usepackage{booktabs,tablefootnote}
\usepackage{multirow}

\begin{document}

\title{Atoms of Thought: Universal EEG Representation Learning with Microstates}
\pagestyle{plain}

\author{Xinyang Tian}
\affiliation{%
  \institution{Institute for Interdisciplinary Information Sciences, Tsinghua University}
  \city{Beijing}
  \country{China}
}
\authornote{Both authors contributed equally to this research.}
\email{xinyangtian368@gmail.com}
\orcid{0009-0008-3168-523X}

\author{Ruitao Liu}
\affiliation{%
  \institution{Institute for Interdisciplinary Information Sciences, Tsinghua University}
  \city{Beijing}
  \country{China}
}
\email{liurt23@mails.tsinghua.edu.cn}
\authornotemark[1]

\author{Ziyi Ye}
\affiliation{%
  \institution{Institute of Trustworthy Embodied AI, Fudan University}
  \city{Shanghai}
  \country{China}
}
\email{zyye@fudan.edu.cn}
\authornote{Research was conducted as a Ph.D. student at Tsinghua University.}

\author{Siyang Xue}
\affiliation{%
  \institution{School of Clinical Medicine, Tsinghua University}
  \city{Beijing}
  \country{China}
}
\email{xuesy22@mails.tsinghua.edu.cn}

\author{Xin Wang}
\affiliation{%
 \institution{Beijing Five Seasons Medical Technology Co., Ltd.}
 \city{Beijing}
 \country{China}
}
\email{wwwwangshin15@gmail.com}

\author{Xuesong Chen}
\affiliation{%
  \institution{Beijing Five Seasons Medical Technology Co., Ltd.}
  \city{Beijing}
  \country{China}
}
\email{chenxuesong1128@163.com}
\authornote{Corresponding author. Email: chenxuesong1128@163.com}
\renewcommand{\shortauthors}{Xinyang Tian et al.}

\begin{abstract}
Learning universal representations from electroencephalogram (EEG) signals is a cutting-edge approach in the field of neuroinformatics and brain-computer interfaces (BCIs). 
Conventionally, EEG is treated as a multivariate temporal signal, where time- or frequency-domain features are extracted for representation learning. 
This paper investigates a simple yet effective EEG representation, i.e., microstates. 
Microstates represent the building blocks of brain activity patterns at a microscopic time scale. 
We build a universal microstate tokenizer from a large medical EEG dataset by clustering continuous EEG signals into sequences of discrete microstates. 
The microstate tokenizer is then adopted universally across a series of downstream tasks, including sleep staging, emotion recognition, and motor imagery classification. 
Experimental results show that EEG representation learning with microstates outperforms traditional time-domain and frequency-domain features under different models and across different tasks. 
Further analysis shows that microstates offer greater interpretability and scalability, thereby opening up applications in both cognitive neuroscience and clinical research.
\end{abstract}


\maketitle
\begin{center}
\small
Accepted by the 3rd International Workshop on Multimodal and Responsible Affective Computing (MRAC 2025).
Version of Record DOI: 10.1145/3746270.3760230.
\end{center}

\section{Introduction}

Electroencephalogram (EEG) signals provide valuable insights into brain activity, making them indispensable in fields such as clinical medicine, neuroscience, and cognitive psychology~\cite{zhou2024interpretablerobustaieeg}. 
For example, EEG has been widely used in clinical settings to detect certain diseases and anomalies~\cite{SONG2004148,Lin2006ANA,Bai2007TheSE}, to investigate neural mechanisms underlying cognitive processes~\cite{Lu2005EEGMA,SEITZMAN2017533}, and to design brain-computer interfaces~\cite{9328561}.
% Accurate interpretation of EEG data is critical for diagnosing medical conditions, understanding brain functions, and creating assistive technologies.
Recently, with the maturity of deep learning techniques, integrating AI technology with EEG analysis has become the new paradigm, significantly improving classification performance in downstream tasks \cite{AhmedtAristizabal2019NeuralMN,afzal2024rest}.

\begin{figure*}[t]
    \centering
    \includegraphics[width=\linewidth]{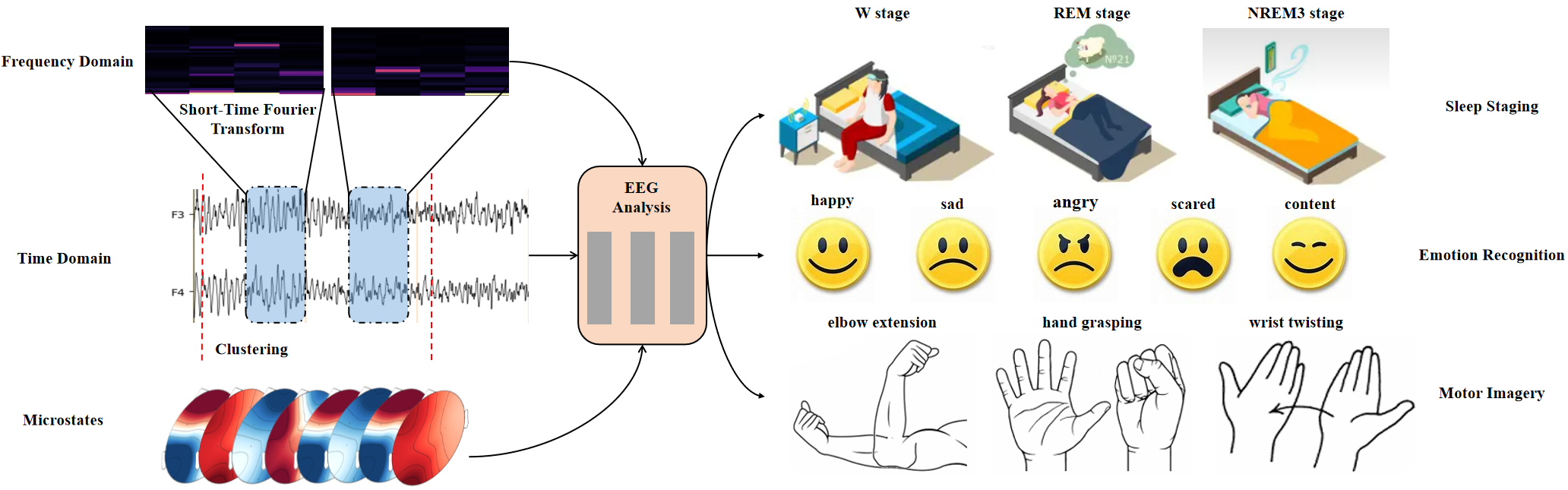}
    \caption{\textbf{Visualization of Different Representations and Downstream Tasks.} Conventional representations mainly reside in the time domain and frequency domain. We propose the microstate representation, which is a universal representation that outperforms other representations under different model structures and across different tasks \cite{lemon,8qw6-f578-23,lee2020motorimageryclassificationsinglearm}.}
    \label{fig:introduction}\Description{Visualization comparing conventional time-domain and frequency-domain representations with the proposed microstate representation, highlighting its superior performance across tasks and models.}
\end{figure*}

Despite these merits, EEG signals are highly non-linear and non-stationary \cite{Subha2010EEGSA}, which pose challenges to extracting effective representations from EEG signals. 
Conventionally, EEG is treated as multivariate time series data with features extracted in the time and frequency domain for further analysis \cite{Subha2010EEGSA}.
Such features come with two major drawbacks.
On the one hand, they \textbf{are susceptible to artifacts and are prone to be confined within a task-specific and subject-specific representation space}.  
Conventional time and frequency domain EEG representations will inevitably incorporate artifacts~\cite{9748967} related to eye blinks, myoelectricity, and the environment. 
Additionally, time and frequency domain EEG representations vary significantly across subjects and tasks, making it challenging to generalize.
% that are specific to a certain person under certain contexts. 
% However, an individual’s EEG signals still possess consistent and meaningful patterns that reflect universal properties despite the artifacts. 
This results in suboptimal performance on a single task and degraded generalizability across different tasks \cite{HU2021177}, and requires huge amounts of task-specific data, which are usually unavailable. 
On the other hand, time- and frequency-domain features are \textbf{unable to uncover transient and dynamic information}. 
Conventional methods often struggle to capture high-resolution EEG features. 
Time-domain information, which directly utilizes raw EEG signals, is often considered inefficient due to its low signal-to-noise ratio (SNR) \cite{wagh2022evaluatinglatentspacerobustness}. 
Frequency-domain information uses a fixed window length, which consequently obscures temporal resolution and results in a certain degree of information loss \cite{Subha2010EEGSA, Cheng2024EEGbasedER}.

To address these challenges, we introduce a novel approach that integrates deep learning with a biologically grounded concept in EEG analysis: EEG microstates \cite{LEHMANN1987271}. 
EEG microstates are quasi-stable discrete patterns of scalp electrical potential that last for brief periods, typically $60$-$120$ milliseconds \cite{MICHEL2018577}. 
While conventional features tend to ignore the physiological and clinical context of EEG signals, EEG microstates are believed to correspond to fundamental and stable cognitive states \cite{EFRON197057,MICHEL2018577,Zechuan2024}. 
Previous researchers have revealed a series of underlying mechanisms of thought and cognition \cite{LEHMANN19981,BRITZ20101162,MILZ2016643,SEITZMAN2017533}. 
Building on such results, we leverage EEG microstates as a discrete and intrinsic representation of brain activity that is more aligned with the underlying neural mechanisms, improving both interpretability and robustness. 
% This method enhances the robustness and stability of downstream tasks, providing a clearer framework for understanding brain dynamics. 
We validate the effectiveness of EEG microstates across three critical tasks---sleep staging, emotion recognition, and motor imagery \cite{9626707,zhou2024interpretablerobustaieeg} and with different models, showing superior performance compared to conventional representations. 
Moreover, we test the accuracy of EEG representation learning with increasing data size, observing that EEG microstates show greater performance gain than conventional features. 
Furthermore, we investigate the distribution of EEG microstates across various cognitive functions and present a potential relationship to interpret cognitive functions with EEG microstates.

The main contributions of this work are as follows:
\begin{itemize}
\item We introduce EEG microstates as a universal representation of brain activity, bridging the gap between deep learning techniques and neural activity patterns. 
\item We demonstrate the effectiveness of this microstate-based approach in three critical tasks---sleep staging, emotion recognition, and motor imagery classification, and with different model structures. Experimental results indicate that the microstate tokenizer initialized in one task can be generalized to a series of downstream tasks, showcasing its universal applicability and alleviating the impact of data scarcity.
\item We conduct in-depth analysis showing that EEG microstate is more scalable than time-domain and frequency-domain methods and can serve as an explainable feature linking to various cognitive functions. 
\end{itemize}

\section{Related Work}
EEG analysis has long been a critical tool in both clinical diagnosis and research, with various representation learning methods to enhance the accuracy of diagnosis. This section elaborates on a variety of techniques developed to extract meaningful information from the brain's electrical activity, particularly in the medical and deep learning fields.

\subsection{EEG Microstates in Cognitive Neuroscience}
EEG microstate analysis was first introduced by \citet{LEHMANN1987271}, and has gained significant attention as a promising tool for representing brief, stable patterns of brain activity. Microstates are thought to reflect fundamental cognitive states that the brain switches between, providing valuable insights into the temporal organization of brain function \cite{MICHEL2018577}. Studies have shown that various diseases, such as epilepsy, sleep disorder and Alzheimer’s disease, can alter EEG microstates \cite{BRODBECK20122129,Tait2019EEGMC,Liu2021AlteredPE,FANG2024109729,piggy,SA2024108266}. Recent research has applied microstate analysis to a wide range of cognitive tasks, including emotion, attention, and social abilities \cite{sheeepy,candy,sheepy,EskandariNasab2024AGM,PMID:39402190}, demonstrating the effectiveness of microstates in understanding cognitive and pathological states.

The most common approach to producing microstates originates from \citet{391164}. They used the k-means clustering method to conduct the EEG microstate analysis, which further become the most popular technique for microstate classification. Other studies have introduced alternative methods for microstate analysis, which are based on a series of clustering algorithms \cite{MAKEIG2004204,Lucia2007SingleSE,article, Pourtois2008BeyondCE,HADRICHE2013448}. 
% Our work adopts this standard clustering approach as well.

However, most existing research has focused on interpreting microstates based on the physical conditions of subjects, while efforts to learn EEG representations for downstream classification and detection tasks remain limited. 
Moreover, the interpretability of microstates and their connection to fundamental cognitive states make them a promising candidate for representing EEG signals in contemporary deep-learning models, yet no current studies have tested this potential.

\subsection{Representation Learning for EEG Analysis}
Machine learning, especially deep learning techniques have been increasingly integrated into EEG analysis to improve the accuracy and efficiency of EEG-based classification tasks. Typically, machine learning models require EEG representations extracted from the raw signals as input, which can be broadly categorized into time- and frequency-domain features.
On the one hand, raw EEG itself can serve as the most straightforward time-domain representation. \citet{alhussaini2019sleeperinterpretablesleepstaging} used fixed-length windows of $30$s segmented from raw EEG signals during prototype learning for sleep staging, which treated the signals as multivariate time series data. \citet{Perslev2021USleepRH} also used raw EEG signals as their CNN-based model representation for sleep staging.

On the other hand, information in the frequency domain is also commonly extracted as EEG representations. 
\citet{V2022107867} used multivariate variational mode decomposition (MVMD) to extract spectral information for emotion recognition.
\citet{9536024} used the Hilbert-Huang transform to analyze scalp EEG signals. 
It has been shown in \cite{Wang2011EEGBasedER,EmotionRecognitionEMD} that using frequency-domain information improves performance in emotion recognition.

Despite the above achievements brought about by deep learning, conventional representations often contain person- or task-specific artifacts \cite{Song2021VariationalIG,wang2024generalizablesleepstagingmultilevel,Wang2024DMMRCD}. Due to the models' susceptibility to noise and artifacts, training such models either undermines their performance and generalizability, or requires a huge amount of person- or task-specific data.

To address these challenges, \citet{afzal2024rest} proposed a novel graphical representation of raw EEG data, which improves seizure detection but is still task-specific. Based on the development of time-domain representations and suitable model structures \cite{Wu2022TimesNetT2,Nie2022ATS,NEURIPS2023_5f9bfdfe} and inspired by the development in natural language processing (NLP), \citet{pmlr-v235-gui24a} proposed a vector quantization pre-training method to obtain representations for downstream tasks. 
\citet{wang2024eegpt} also utilized a pre-training paradigm to extract relevant representations by spatio-temporal representation alignment in order to depict the brain. They observed that such representations can be better generalized across downstream tasks, but consume a large amount of computational power and time. Moreover, the input EEG signal of the pre-trained model is still treated as multivariate time series data.

\section{EEG Representations}
This section lists conventional EEG representations in the time-domain and frequency-domain, and our microstate representation. It also elaborate on detailed methods and procedures to construct different representations.

\subsection{Problem formulation}
The objective of EEG signal analysis and physical state prediction can be defined as follows: 
We are given the input raw EEG signal $s\in\mathbb{R}^{C\times f_sT}$ where $C$ denotes the number of channels, $f_s$ is the sampling frequency and $T$ is the sample duration. 
The EEG signal analysis aims to predict the physical state of the sampled subject, which can be represented by a sequence of discrete labels $l=L^{f_lT}$ where $L=\{a_1,a_2,\ldots,a_m\}$ is the set of labels and $f_l$ is the state frequency.

\subsection{Time-Domain Features Extraction}
The most straightforward approach for EEG analysis is to directly utilize the time-domain information features, i.e., the raw EEG signals \cite{Subha2010EEGSA}. In this setting, the raw EEG signal is sliced into fixed-length windows with duration $T_w$, and the feature $s_{time,w}$ will be of the shape of $\mathbb{R}^{C\times f_sT_w}$, with the corresponding labels $l_w\in L^{f_lT_w}$.

\subsection{Frequency-Domain Features Extraction}
Raw EEG signals often obscure frequency information, and thus, sometimes directly using it does not produce desirable results. 
Therefore, a common approach to handling such time-domain signals is to use their corresponding frequency-domain signals as features \cite{doi:10.1504/IJBET.2017.082661}.

\paragraph{Frequency Bands.}The frequency-domain representations are extracted based on the frequency power distribution among frequency bands\cite{xiao20214dattentionbasedneuralnetwork}.
The frequency domain are divided into several frequency bands, including the $\delta$-band ($0.5\sim4$Hz), $\theta$-band ($4\sim8$Hz), $\alpha$-band ($8\sim12$Hz), $\sigma$-band ($12\sim16$Hz), $\beta$-band ($16\sim30$Hz) and $\gamma$-band ($30\sim40$Hz). 

\paragraph{Short-Time Fourier Transform (STFT)}Time-frequency transformation can be carried out via numerous methods, namely short-time Fourier transform (STFT), discrete/continuous wavelet transform (DWT/CWT), and empirical mode decomposition (EMD) \cite{AKAN2021103216,KHARE2024102019}. Short-time Fourier transform, owing to its straightforwardness and thorough theoretical analysis, is applied in many EEG-related tasks \cite{WANG2020107506,Hwang20201323,9154557,10041186}. Consequently, we choose this method as our frequency-domain baseline.

Given the raw EEG signal of a single measurement channel $s_{sin}\in\mathbb{R}^{f_sT}$, we perform short-time Fourier transform with fixed window size $t_w$ and overlap ratio $r_o$. The length after the short-time Fourier transform will be
\begin{equation*}
    l_{freq}=\left[\frac{f_sT-f_st_w}{(1-r_o)f_st_w}\right]+1
\end{equation*}
and if we leave out the margin then the resulting length will be
\begin{equation*}
    \begin{aligned}
    l_{freq}&=\frac{f_sT}{(1-r_o)f_st_w}=f_{freq}T\\
    f_{freq}&=\frac{1}{(1-r_o)t_w}
    \end{aligned}
\end{equation*}

Using STFT, we obtain a frequency axis $F_a$ and time axis $T_a$ and the amplitude of the signal at each frequency $f\in F_a$ and time point $t\in T_a$. Note that $|T_a|=l'$ and hence the output shape is $s_{f,t,sin}\in\mathbb{R}^{|F_a|\times l'}$.

\begin{figure*}[t]
    \centering
    \includegraphics[width=\linewidth]{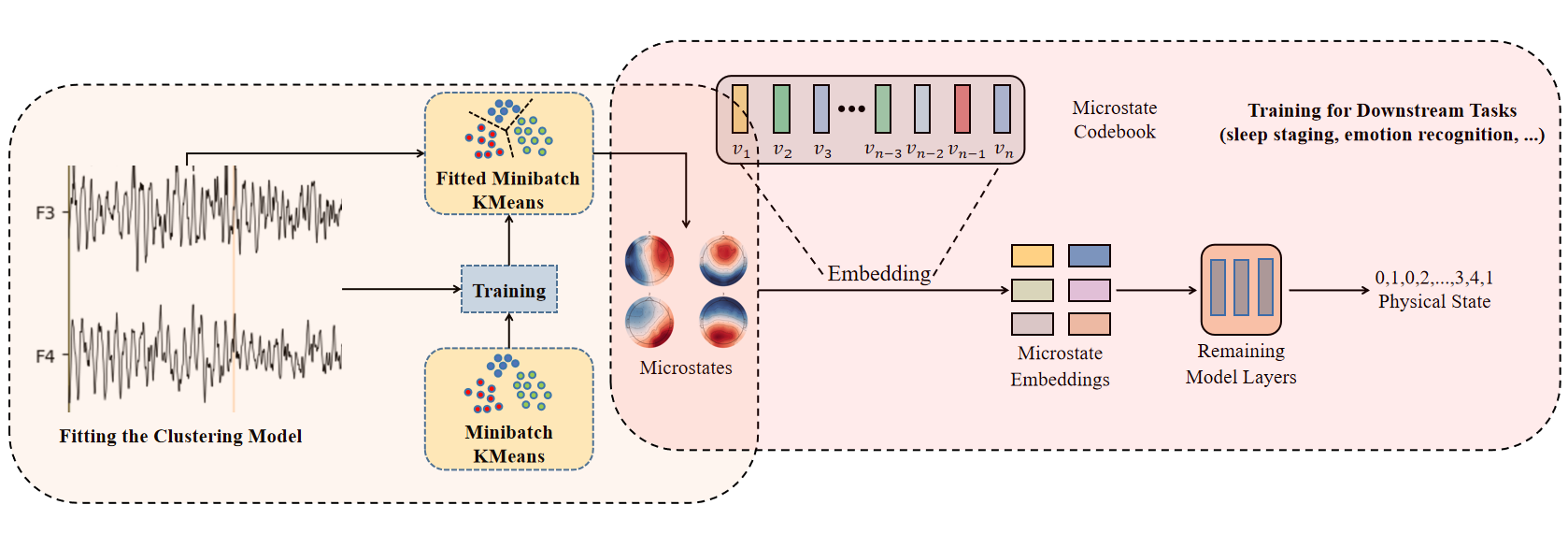}
    \caption{\textbf{Pipeline of our Work.} Our work can be broken into two parts. The first involves fitting a tokenizer to extract microstates from six EEG channels F$3$, F$4$, C$3$, C$4$, O$1$, O$2$. The second consists of training different models on the microstate signals and performing downstream tasks. The tokenizer of the first stage is independent of the models of the second stage. For each task, the model always includes an embedding layer to convert the discrete microstates into high-dimensional embeddings~\cite{lemon}.}
    \label{fig:pipeline}\Description{Own pipeline.}
\end{figure*}

\paragraph{Band Power Integration.}Now that we have obtained the single channel data $s_{freq,sin}\in\mathbb{R}^{F\times l_{freq}T}$ where $F$ is the frequency resolution, we integrate the rows that correspond to each frequency band to obtain the total power within that band. In this case, the integration result will have shape $\mathbb{R}^{B\times l_{freq}T}$. By flattening and stacking all channels, the final result has shape $s_{freq}\in\mathbb{R}^{N\times Bf_{freq}T}$.

\subsection{Unsupervised Microstate Tokenizer}
Our clustering method is based on k-means \cite{391164}. But to guarantee that the clustering model has a sufficient level of generalizability, unlike previous works, we have to fit the clustering model on a huge amount of data. To allow the model to take all data points into consideration without consuming too much memory, we use incremental learning by dividing data into small batches of size $n$.

\subsubsection{Stream Clustering~\cite{AGGARWAL200381}}
The original k-means algorithm clusters the data points by initially selecting $k$ centers, grouping data points according to their distance to the centers, and computing the centroid of each cluster as the new centers. The algorithm terminates when reaching the maximum iterations or the positions of the centers have converged.

Streaming k-means follows a fashion similar to classical k-means, but each time it uses only a small batch of data to update the cluster centers. Therefore, we do not need to store the entire dataset in memory, while only need to store a batch. This renders better performance than using only a small portion of data, and reduces memory consumption compared to clustering on all data points. The procedure is shown in Algorithm \hyperref[algorithm:stream_clustering]{1}.

\begin{algorithm}\label{algorithm:stream_clustering}
    \caption{Streaming K-Means}
    \begin{algorithmic}
    \STATE Initialize cluster centers $c_1,c_2,\ldots,c_k\in\mathbb{R}^C$
    \STATE $iter\leftarrow0$
    \WHILE{$iter<max\_iter$ and centers have not converged}
        \STATE Get a new batch $d_1,d_2,\ldots,d_n\in\mathbb{R}^C$
        \STATE $S_i\leftarrow\{d_j|\arg\min_t\|d_j-c_t\|^2=i\}$
        \STATE $c_i\leftarrow\frac{1}{|S_i|}\sum_{r=1}^{|S_i|}{S_i}_r$
    \ENDWHILE
    \end{algorithmic}
\end{algorithm}

\subsubsection{Fitting the Clustering Model}

We adopt the following experimental setup.

\paragraph{Dataset.}We select the \textbf{Human Sleep Project (HSP)}\label{data:hsp} dataset to fit the clustering model \cite{Westover2023}. The dataset includes polysomnography (PSG) data from over $20$K subjects and is still growing in size. The reasons why we use EEG data recorded during sleep are as follows:
\begin{itemize}
    \item It is usually difficult and costly to record EEG signals during wakefulness, and these datasets are usually limited in size
    \item Currently the community has abundant sleep data. They can also reflect the cognitive level and consciousness, despite the fact that their being less organized, noisy and has a limited number of channels (usually $6$ channels)
    \item We want to test whether we can model brain activity during sleep data, which can be generalized to downstream tasks during wakefulness
\end{itemize}

\paragraph{Extracting target channels.}Since PSG signal involves numerous components like EEG, EOG, and ECG, the number of channels comes with varying sizes due to loss of data or shortage of equipment. Consequently, we filter out all channels except $N$ of them that are present in all samples. The generic method will be clustering the data by treating them as $N$ dimensional points. In the HSP dataset, only the channels F$3$, F$4$, C$3$, C$4$, O$1$, O$2$ are present across a relatively large amount of subjects, whereas other channels only appears sporadically among very limited number of subjects. Consequently these $6$ leads are selected for clustering, since we have to extract a generalizable representation across subjects in order to obtain a truly universal representation. After extracting the necessary channels from the original data $s\in\mathbb{R}^{C\times f_sT}$ we obtain the filtered data $s_{ext}\in\mathbb{R}^{N\times f_sT}$ where $N=6$.

\begin{table*}[t]
\caption{Classification accuracies and model parameters for sleep staging under different representations and different model architectures on the Human Sleep Project (HSP) dataset. 
The highest performance among all representations under a certain model is highlighted in \textbf{boldface}.}
\label{table:sleep_staging_results}
\centering
\begin{small}
\begin{sc}
\begin{tabular}{lccccccccc}
\toprule
Representation & \multicolumn{3}{c}{CNN+LSTM} & \multicolumn{3}{c}{Sleep Transformer} & \multicolumn{3}{c}{Sleep Net Zero} \\
               & Acc & Kappa & params & Acc & Kappa & params & Acc & Kappa & params \\
\midrule
Raw EEG (Time Domain)   & $0.710$ & $0.597$ & $707$K & $0.786$ & $0.702$ & $3.2$M & $0.793$ & $0.713$ & $10.9$M\tablefootnote{For Sleep Net Zero, more parameters are used since the input size of raw EEG signals is $(6,30000)$ which is six times that of microstates $(30000,6)$ and $17$ times that of frequency-domain signals $(6,1800)$.} \\
STFT (Frequency Domain) & $0.778$ & $0.690$ & $692$K & $0.790$ & $0.710$ & $3.2$M & $0.794$ & $0.711$ & $3.2$M \\
Microstates (Ours)      & $\boldsymbol{0.801}$ & $\boldsymbol{0.722}$ & $687$K & $\boldsymbol{0.810}$ & $\boldsymbol{0.736}$ & $3.4$M & $\boldsymbol{0.810}$ & $\boldsymbol{0.736}$ & $3.2$M \\
\bottomrule
\end{tabular}
\end{sc}
\end{small}
\end{table*}

\paragraph{Filtering.}A bandpass filter with low pass $1$Hz and high pass $40$Hz is applied to retain the most relevant frequency bands (e.g., delta, theta, alpha, beta). The array shape remains unchanged during this operation.

\paragraph{Resampling.}The HSP dataset comes with different sample frequencies, including $256$Hz and $512$Hz. We resample the signals to $100$Hz for better clustering results. Now the data shape becomes $s_{res}\in\mathbb{R}^{N\times f_{res}T}$ with $f_{res}=100$Hz.

\paragraph{Global field power (GFP) peaks extraction.}Global field power is computed as the standard deviation of all sensors. The peaks are defined as their local maxima, which have the highest signal-to-noise ratio \cite{MICHEL2018577}. Therefore, we extract GFP peaks on the six channels. This produces the final input for the clustering model, which has size $s_{gfp}\in\mathbb{R}^{N\times t}$ where $t$ denotes the number of GFP peaks in the sequence $\mathbb{R}^{N\times f_{res}T}$.

\paragraph{Fitting the clustering model.}After obtaining the data of shape $s_{gfp}\in\mathbb{R}^{N\times t}$, we set the number of clusters $k=1000$ and $n=50$ for batch size and fit the GFP peaks.

\subsubsection{Constructing the Microstates}
Having fitted clustering model, we can apply it to raw EEG signals of shape $s\in\mathbb{R}^{N\times f_sT}$ to obtain the microstate sequence $c\in S^{f_sT}, S\in\{b_1,b_2,\ldots,b_k\}$ with $k=1000$. $S$ is the set of microstates.

\section{Downstream Tasks}
In this section, we introduce the experimental setup for testing the performance of our microstate representation and conventional representations. The microstate sequence is produced by the tokenizer trained in the previous section (the fitted KMeans in Figure~\ref{fig:pipeline}).

\subsection{Sleep Staging}
During different sleep stages, brain activity varies accordingly. 
This lies the foundation for predicting sleep stages using EEG signals.

\paragraph{Dataset.}For sleep staging, we use the HSP dataset \cite{Westover2023} mentioned in \hyperref[data:hsp]{\textit{fitting the clustering model}} to test the performance of our representation.

\paragraph{Sleep stages.}For sleep staging, $L$ consists of the five sleep stages: \{W,N$1$,N$2$,N$3$,R\}. W corresponds to wake stage, N$1$, N$2$ and N$3$ correspond to different non-rapid eye movement (NREM) stages, and R corresponds to rapid eye movement (REM) stage.

\paragraph{Preprocessing.}The EEG signal is filtered and resampled to $f_{res}=100$Hz. For extracting frequency-domain information, we choose $t_w=1$s and $r_o=0$. Finally, the EEG signal is slices into $T_w=300$s windows.

\paragraph{Model Architecture.}To test the universality of our microstate representation, we adopt the following model structures:

\textit{CNN+LSTM.}\cite{Shao2022AHD} This model architecture consists of $3$ CNNs and $2$ GRUs for extracting spatial and temporal information, and fully connected layers for classification. An embedding layer is added for our microstate representation.

\textit{Sleep Transformer.}\cite{9697331} Sleep Transformer uses $2$ Roformer \cite{su2023roformerenhancedtransformerrotary} layers to extract the local and global features before inputting into the final linear layer. An embedding layer and $2$ CNNs are added for our microstate representation.

\textit{Sleep Net Zero.}\cite{li2024sleepnetzero} The model consists of a feature extraction unit composed of several residual blocks, a Roformer layer and a linear layer. To adapt to our microstate representation, we add an embedding layer and $4$ CNNs, and remove the feature extraction unit.

For time- and frequency-domain signals, we change the embedding layer into a convolution layer which functions similarly as the embedding.

\paragraph{Loss Function.}The loss function is set as cross-entropy loss. Suppose that the output of the fully connected layer is $(h_1,h_2,h_3,h_4,h_5)$, which are the scores of the five classes. We perform a softmax on the scores and the cross entropy loss is defined as follows:
\begin{equation*}
    \mbox{loss}=-\sum_{i=1}^5p(i)\log\mbox{Softmax}(h_i)
\end{equation*}
which is minimized when the correct label $j$ has score $h_j$ significantly larger than the other labels.

\subsection{Emotion Recognition}
Emotions are a key part of our physical state, and have a strong connection with brain activity. Consequently, our work involves training a microstate-based model for emotion classification.

\paragraph{Dataset.}We use the \textbf{SEED} dataset \cite{duan2013differential,zheng2015investigating} for emotion recognition. The SEED dataset consists of $15$ subjects watching video clips that express different emotions. The overall tone is categorized as positive, negative, and neutral. The EEG signal is recorded with $62$ channels and at a frequency $200$Hz.

\paragraph{Emotion Labels.}We utilize the overall tone of each movie clip as our labels, and thus $L$ consists of positive, negative, and neutral.

\paragraph{Preprocessing.}We directly use the sample frequency $f_s=200$Hz. We pad all samples to $T_w=265$s. Other configurations are the same as in sleep staging.

\paragraph{Model Architecture.}The model used here is a CNN-based classifier \cite{seedcnn} with similar modifications and cross-entropy loss. It has $5$ CNNs and $4$ linear layers.

\begin{table}[t]
\caption{Classification accuracies and model parameters for emotion recognition (CNN-based model, SEED dataset) under different representations. The highest performance among all representations is highlighted in \textbf{boldface}.}
\begin{center}
\begin{small}
\begin{sc}
\begin{tabular}{lccc}
\toprule
Representation & Accuracy & Kappa & Params \\
\midrule
Raw EEG (Time Domain) & $0.846$ & $0.769$ & $19.1$M \\
STFT (Frequency Domain) & $0.797$ & $0.694$ & $19.1$M \\
Microstates (Ours)   & $\boldsymbol{0.862}$ & $\boldsymbol{0.793}$ & $20.1$M \\
\bottomrule
\end{tabular}
\end{sc}
\end{small}
\end{center}
\label{table:emotion_recognition_results}
\end{table}

\begin{table}[t]
\caption{Classification accuracies, model parameters for motor imagery classification (ResNet, Motor Movement/Imagery dataset) under different representations. The highest performance among all representations is highlighted in \textbf{boldface}.}
\begin{center}
\begin{small}
\begin{sc}
\begin{tabular}{lccc}
\toprule
Representation & Accuracy & Kappa & Params \\
\midrule
Raw EEG (Time Domain) & $0.362$ & $0.149$ & $20.3$M \\
STFT (Frequency Domain) & $0.323$ & $0.097$ & $21.5$M \\
Microstates (Ours)   & $\boldsymbol{0.437}$ & $\boldsymbol{0.250}$ & $21.4$M \\
\bottomrule
\end{tabular}
\end{sc}
\end{small}
\end{center}
\label{table:motor_imagery_classification_results}
\end{table}

\subsection{Motor Imagery Classification}
Physical movement or imagination is another important factor of human physical status. Therefore our work involves predicting the movement or imagination activity via microstate sequences.

\paragraph{Dataset.}We use the \textbf{Motor Movement/Imagery Dataset} \cite{goldberger2000physionet,1300799} for the task of motor imagery classification. The dataset consists of EEG signals sampled from $109$ subjects. Each subject underwent $14$ trials involving four tasks and two baseline rest sessions. The tasks are as follows:
\begin{itemize}
\item Task $1$: Open and close the left or right fist.
\item Task $2$: Imagine opening and closing the left or right fist.
\item Task $3$: Open and close both fists or both feet.
\item Task $4$: Imagine opening and closing both fists or feet.
\end{itemize}

\paragraph{Movement/Imagery Labels.}In this setting, we focus on the onset of movement/imagery and the labels $L$ consists of left hand, right hand, both hands and both feet.

\paragraph{Preprocessing.}In this dataset, each label corresponds to roughly $4$s of EEG signals, and hence we choose $T_w=4$s. Other configurations are the same as the above experiments.

\paragraph{Model Architecture.}The model architecture is based on \cite{Cheng2020SubjectAwareCL}, which consists of a CNN and $3$ residual blocks followed by $4$ linear layers. We use configurations similar to the above.

\section{Results and Analysis}
We compared our proposed representation with conventional time- and frequency-domain representations across different tasks and under different model configurations.

\subsection{Evaluation Metrics}
We evaluated our representation with different model structures and across different tasks. The evaluation metrics are the classification accuracy and Cohen's Kappa.

\subsection{Microstates as a Universal Representation}
We compared the accuracy and Cohen's Kappa on the test set under $3$ different models and across three key tasks with different representations. 
Results on sleep staging are shown in Table~\ref{table:sleep_staging_results}, and results on emotion recognition and motor imagery classification are shown in  Table~\ref{table:emotion_recognition_results} and Table~\ref{table:motor_imagery_classification_results}, respectively.

\begin{figure}[htbp]
    \centering
    \begin{minipage}{0.45\linewidth}
        \centering
        \includegraphics[width=\linewidth]{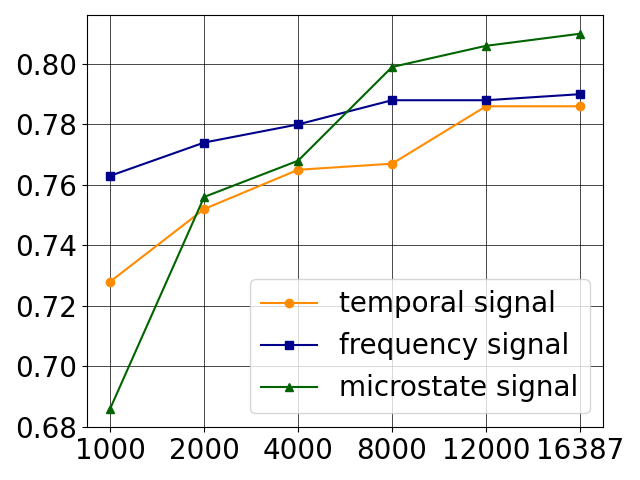}
        \text{(a) Accuracy}
    \end{minipage}
    \hspace{0.2cm}
    \begin{minipage}{0.45\linewidth}
        \centering
        \includegraphics[width=\linewidth]{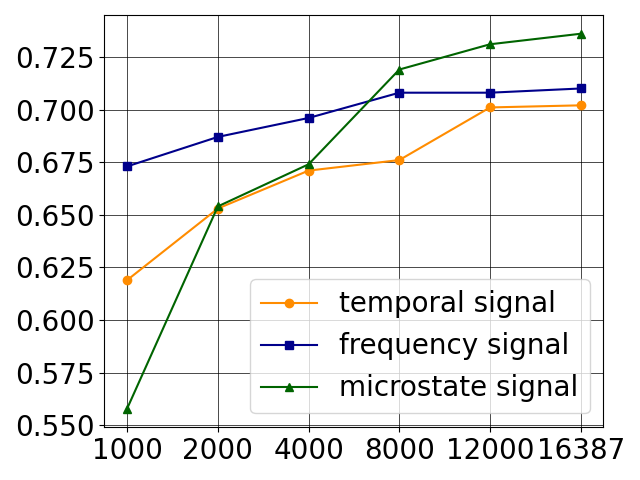}
        \text{(b) Cohen's Kappa}
    \end{minipage}
    \captionsetup{belowskip=0pt}
    \caption{Accuracy (left) and Cohen's Kappa (right) with different representations under Sleep Transformer and different number of samples.}
    \label{figure:small_sample_sleep_staging}\Description{Testing on a smaller number of data.}
\end{figure}

From Table~\ref{table:sleep_staging_results}, we observe that EEG representation with microstates outperforms time domain and frequency features in three different backbone models, including CNN+LSTM, Sleep Transformer, and Sleep Net Zero.
Among the results, microstates achieve the highest accuracy of $0.81$ using a sleep transformer or sleep net zero.
This indicates that EEG microstates have the potential to serve as a universal representation and outperform temporal- and frequency-domain features across tasks and model structures.
Similar observations are obtained in the emotion recognition task based on a CNN-based model and on the motor imagery classification task based on a ResNet model.

We also record the standard deviation of the performance of microstate representation on Sleep-Net-Zero, which gives an accuracy of $0.808(\pm1.897\cdot10^{-3})$ and Kappa $0.733(\pm2.482\cdot10^{-3})$.

% achieving the best performance across tasks and model structures. 
We further compare the performance of time- and frequency-domain features with microstates. 
We see that frequency-domain representation performs well on sleep staging, while producing suboptimal results on other tasks. 
We suspect that this is because sleep staging is highly frequency-associated, while on other tasks, frequency-domain features may be weak due to information loss \cite{Subha2010EEGSA, Cheng2024EEGbasedER} in raw EEG signals. 
On the other hand, raw EEG signals are often subject to noise \cite{wagh2022evaluatinglatentspacerobustness} and does not produce optimal results.
Compared to time- and frequency-domain features, microstates present a robust performance across datasets and classification models.
This demonstrates that the microstate representation obtained from sleep EEG data can be generalized to various critical tasks and different models, serving as a universal representation.

\begin{table}[t]
\caption{Classification accuracies and model parameters for emotion recognition (CNN-based model, SEED dataset) including full channels ($62$ in total). The highest performance among all representations is highlighted in \textbf{boldface}.}
\begin{center}
\begin{small}
\begin{sc}
\begin{tabular}{lccc}
\toprule
Representation & Accuracy & Kappa & Params \\
\midrule
Raw EEG ($6$ channels) & $0.846$ & $0.769$ & $19.1$M \\
Raw EEG ($62$ channels) & $0.854$ & $0.778$ & $19.2$M \\
STFT ($6$ channels) & $0.797$ & $0.694$ & $19.1$M \\
STFT ($62$ channels) & $0.854$ & $0.778$ & $19.1$M\\
Microstates (Ours)   & $\boldsymbol{0.862}$ & $\boldsymbol{0.793}$ & $20.1$M \\
\bottomrule
\end{tabular}
\end{sc}
\end{small}
\end{center}
\label{table:full_channel_results}
\end{table}

\begin{figure}[htbp]
    \centering
    \includegraphics[width=0.9\linewidth]{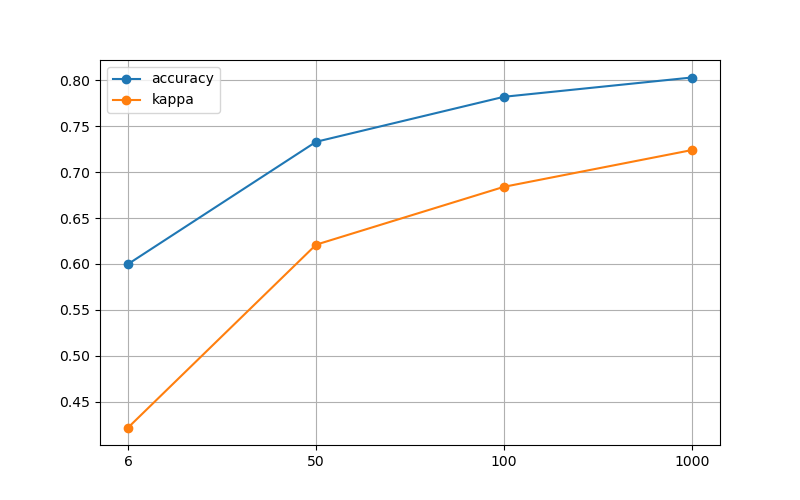}
    \captionsetup{belowskip=0pt}
    \caption{Accuracy and Cohen's Kappa under Sleep Net Zero with different number of microstates.}
    \label{figure:class_sleep_staging}
    \Description{Testing accuracy and kappa with different classes.}
\end{figure}

\subsection{Results using Full Channel Data}
Due to data constraint, the microstate tokenizer is trained on data from only $6$ channels. To allow for a comprehensive comparison, we test the performance of the CNN classifier on the SEED dataset using full channels ($62$ channels in total). The results are shown in Table~\ref{table:full_channel_results}.

From the results we see that using EEG signals from the $6$ channels can achieve similar results to that of using full data, as is seen from the raw EEG signals that increasing the number of channels does not significantly boost performance. Notice that the performance of frequency-domain representation increases significantly, which we conjecture that it is because the additional channels compensate for the information loss during the time-frequency transformation. Results show that using only $6$ channels does not significantly degrade performance, which justifies our clustering on these channels.

\begin{figure*}[t]
    \centering
    \includegraphics[width=\linewidth]{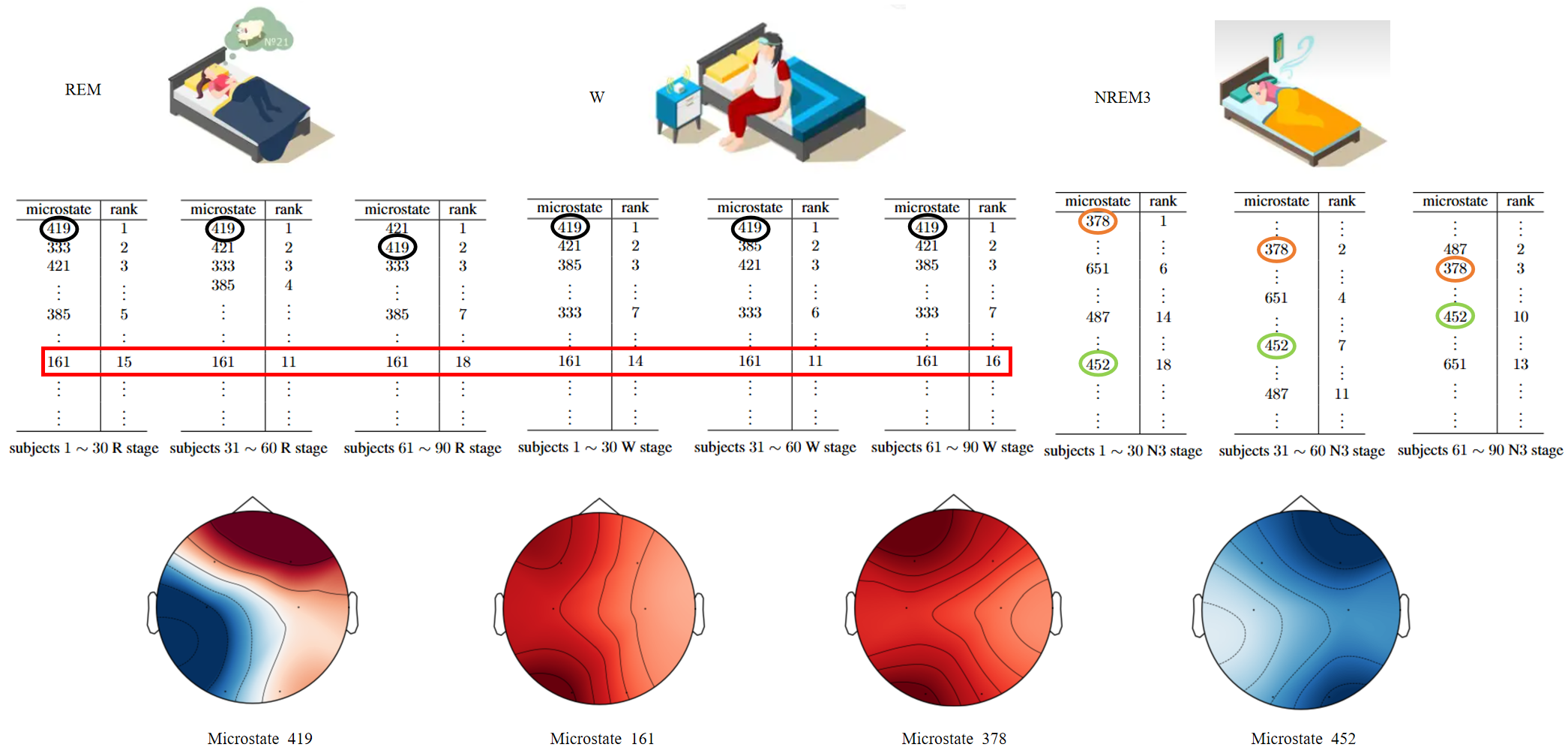}
    \caption{\textbf{Visualizing Microstates Distribution.} Visualization of the distribution of different microstates among different subjects undergoing different sleep stages. We can see that the microstate representation simutaneously retain the similarity between subjects and between W and R stage, while also preserves the difference between W stage N$3$ stage.}
    \label{fig:interpretation}\Description{Visualization different microstates and their meanings.}
\end{figure*}

\subsection{Microstates as a Scalable Representation}
We further test the performance of EEG representation learning with microstates across different scales of training data in the sleep staging task under Sleep Transformer on the HSP dataset.
We also find that microstates offer a more scalable representation. 
% We test the performance of sleep staging on different numbers of training samples under Sleep Transformer on the HSP dataset. 
As shown in Figure~\ref{figure:small_sample_sleep_staging}, microstates do not exhibit strengthened performance when the size of the training data is smaller than 2,000.
However, when the number of samples increases, the performance of the microstate representation shows a more pronounced performance gain in comparison to other features. 
Our experiment demonstrates that the microstate representation is also capable of scaling across the size of the training data.
This reveals the potential of EEG representation with microstates, especially using deep learning methods and increased data size.

On the other side, we tested the performance of our tokenizer under different number of clusters by selecting different parameters $k$ for clustering. We evaluated the model performance on the validation set with different number of microstates. The results are shown in Table~\ref{figure:class_sleep_staging}. From the results we see that the performance increases while the number of microstates increases. Results show that the performance of the microstate representation also scales with increasing number of microstates.

\subsection{Interpreting Microstates}
In this section, we give an analysis of the interpretability of the microstate representation, which in turn leads to its better performance over other representations.

As mentioned in \cite{9748967}, one challenge in analyzing EEG signals is that they are highly subject-dependent and vary significantly across different people. Consequently, it is hard to extract effective inter-subject representations using conventional time- or frequency-domain information.
Microstates solve this issue by providing a coarse-grained discrete representation that groups similar EEG states together. 
Its clustering-based nature guarantees its capability to extract universal features while retaining the differences.

We analyze the proportion of the most $20$ frequently-occurring microstates among groups of $30$ subjects under W, N$3$, and R stage. Results in Figure~\ref{fig:interpretation} show that under the same sleep stages, the most frequent microstates are common across all subject groups. For example, the microstates $419,421,385,333$ occur with high frequency among all subject groups during W and R stage, whereas the microstates $487,378,452,651$ occur with high frequency among all groups under N$3$ stage. This suggests that the microstate representation captures the similarity between subjects, albeit their having different EEG voltages. Hence this in turn prevents the model from being distracted towards personal specific nuances.

Furthermore, the most frequent microstates under W and R stages both contain $419$ and $161$. The state $419$ has all its channels below $2.2\mu$V, denoting a state with a weak EEG signal, while the state $161$ has its voltage within the interval $4\sim11\mu$V, which is also relatively low. This is consistent with the fact that during W stage, the EEG signal is dominated by $\alpha$ waves, which have a low amplitude and high frequency. Also, during R stage, the brain activity is similar to W stage since this is when dreams take place \cite{ELHADIRI2024102664}. Consequently, it does not come as a surprise that W and R stages share many microstates in common, indicating a similar brain activity pattern. However, the microstate $378$ denotes signals within the interval $10\sim24\mu$V, and $452$ has signals within $-5\sim-21\mu$V. Both of them are relatively strong brain activity. This is again consistent with the fact that during N$3$ the EEG signal has a larger portion of $\delta$ waves with a larger amplitude \cite{ZHANG2024651}. This suggests that microstates are capable of extracting the similarities between sleep stages, while also retaining their differences. 

\balance
\section{Discussions and Conclusion}
In this work, we introduce EEG microstates as a clinically grounded approach for integrating deep learning and EEG signal analysis. 

Our approach improves the representation of brain activity by aligning more closely with the underlying neural mechanisms and cognitive activities, enhancing both clinical and research applications. 
Experimental results demonstrate the effectiveness of EEG microstates in three critical tasks---sleep staging, emotion recognition, and motor imagery classification and across different models, where it outperforms traditional time-domain and frequency-domain methods. 

Furthermore, we show that EEG microstates present more performance gain than time- and frequency-domain features when scaling the data size, indicating that EEG microstates can alleviate the burden of data scarcity and pave the way to more scalable settings. We also show that EEG microstates can provide interpretable insights for EEG analysis and deep learning, offering a promising direction for future research and clinical practice. The adoption of EEG microstates holds significant potential for advancing both cognitive neuroscience and the field of clinical diagnostics.

Several limitations guide future work, such as:
\begin{itemize}
\item We only experimented with limited tasks and limited number of datasets. Particularly, the training of the tokenizer was only performed on sleep data. This is reasonable because the HSP dataset is the largest, but more research can be conducted across tasks in the future.
\item We only focused on the representation side. Based on the microstate representation, we hypothesize that it is possible to develop a pre-trained model that can generalize to several EEG-related downstream tasks.
\end{itemize}

Above all, we believe that combining deep learning techniques with biologically grounded EEG microstates opens up a portal to future research on improving the accuracy of EEG analysis across different tasks and on uncovering more correlations between microstates and brain activity. Future work might involve reconstructing brain signals with more channels to alleviate the lack of channel data.

\newpage
\bibliographystyle{ACM-Reference-Format}
\balance
\bibliography{sample-base}
\begin{acks}
This work is supported by the Ministry of Science and Technology of China STI2030-Major Projects (No. 2021ZD0201900, 2021ZD0201902).
The computations in this research were performed using the CFFF platform of Fudan University.
\end{acks}

\appendix

\section{Experimental Setup}\label{appendix:A}
This section gives the detailed experimental configuration of our downstream tasks.

\subsection{Sleep Staging}
\subsubsection{Dataset}
The dataset used is the \textbf{Human Sleep Project (HSP)} dataset \cite{Westover2023}. This dataset includes PSG signals from over $20$K subjects. Signals are sampled under various frequencies including $256$Hz and $512$Hz. Each sample includes a night's sleep of a subject sampled, with sleep stages annotated every $30$ seconds. Equivalently, we have the raw EEG signals $s\in\mathbb{R}^{C\times f_sT}$ where $C$ consists of different classes of channels such as EOG, ECG, and EEG and differs across subjects, $f_s$ denotes the sample frequency which also differs across subjects, and $T$ is the time duration of a night's sleep, which is typically $6$-$7$ hours. The label frequency is $f_l=\frac{1}{30}$Hz.

\subsubsection{Preprocessing}
\paragraph{Extracting target channels.}As for constructing the microstates, we have to filter out a fixed number of channels. To achieve this goal, we extract $N=6$ channels which is common among all samples. The EEG leads are shown in the following diagram \cite{duan2013differential,zheng2015investigating}. The channels chosen are F$3$, F$4$, C$3$, C$4$, O$1$, O$2$. After extraction, the EEG signals have shape $s_{ext}\in\mathbb{R}^{N\times f_sT}$.

\begin{figure}[ht]
    \centering
    \includegraphics[width=0.45\linewidth]{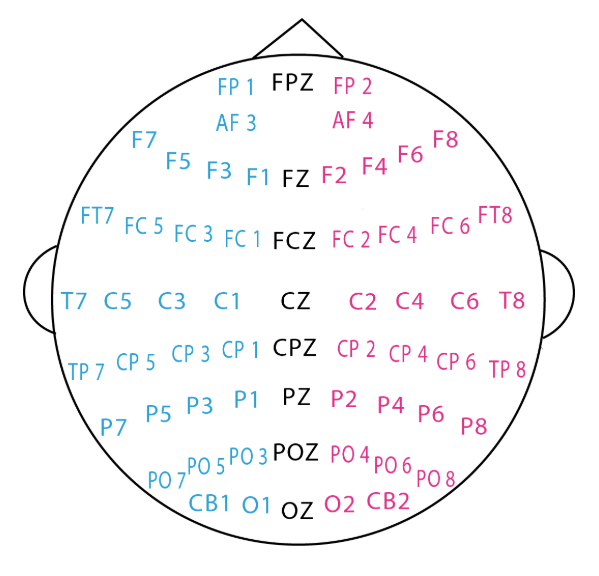}
    \caption{The Distribution of EEG Leads}
    \label{fig:eeg_leads}
\end{figure}

\paragraph{Filtering and Resampling.}The raw EEG signals are then bandpass filtered between $1$Hz and $40$Hz, followed by a resampling at $f_{res}=100$Hz. After these procedures, the raw EEG signals now have shape $s_{res}\in\mathbb{R}^{N\times f_{res}T}$. Having obtained the resampled data, we construct the representations accordingly.

\paragraph{Constructing microstates.}The fitted clustering model is applied to the resampled data $s_{res}\in\mathbb{R}^{N\times f_sT}$. The result is a microstate sequence $c\in S^{f_{res}T}$ where $S\in\{b_1,b_2,\ldots,b_k\}$ is a set of $k$ discrete states. Here we let $k=1000$.

\paragraph{Constructing baseline.}We directly use the raw EEG signals for time-domain features. The input shape is $s_{time}\in\mathbb{R}^{N\times f_{res}T}$.

To extract frequency information, we use short-time Fourier transform. The major goal is to decompose the signal into powers at different frequencies. \cite{AKAN2021103216}. For a given frequency $f$, the power is computed as follows:
\begin{equation*}
    \begin{aligned}
     X(t,f)&=\int_{-\infty}^\infty w(t-\tau)s(\tau)e^{-i2\pi f\tau}d\tau\\
     p(t,f)&=\frac{1}{2\pi}|X(t,f)|^2
    \end{aligned}
\end{equation*}
where $w(t)$ is a window function. Note that
\begin{equation*}
    X'(t,f)=\int_{-\infty}^\infty s(\tau)e^{-i2\pi f\tau}d\tau
\end{equation*}
is the usual Fourier transform, and the window function $w(t)$ only has finite support which serves as a short-time weighted sum of the integral. We use the Hann window function defined as follows:
\begin{equation*}
w(t)=\left\{
    \begin{aligned}
    \frac{1}{2}(1-\cos\frac{2\pi t}{T})&&|t|\le\frac{T}{2}\\
    0&&|t|>\frac{T}{2}
    \end{aligned}
\right.
\end{equation*}
where $T$ is the window length. In our setting we set $T=t_w=1$s. The overlap ratio is set to be $r_o=0$.

Using the above approach, the processed frequency-domain signals have length $l'$ where
\begin{equation*}
    l'=\left[\frac{f_{res}T-f_{res}t_w}{(1-r_o)f_{res}t_w}\right]+1
\end{equation*}
is the number of windows. Leaving out the margin, we have that
\begin{equation*}
    l'=\frac{T}{(1-r_o)t_w}=f_{freq}T
\end{equation*}
here for $r_o=0$ and $t_w=1$s, we have $f_{freq}=1$Hz.

Having calculated the power $p(t,f)$ at frequency $f$ and time $t$, we obtain the spectrogram $P\in\mathbb{R}^{F\times f_{freq}T}$ for each channel, where $F$ is the frequency axis and $f_{freq}T$ is the time axis.

Next, we apply band integration. Human EEG signal is divided into the following frequency bands:
\begin{itemize}
\item $\delta$-band: $0.5\sim4$Hz
\item $\theta$-band: $4\sim8$Hz
\item $\alpha$-band: $8\sim12$Hz
\item $\sigma$-band: $12\sim16$Hz
\item $\beta$-band: $16\sim30$Hz
\item $\gamma$-band: $30\sim40$Hz
\end{itemize}
and we combine the powers among $F$ within each band. We use simpson integration as our numerical quadrature method, which is defined as
\begin{equation*}
    \int_a^bf(x)dx=\frac{b-a}{6}\left(f(a)+4f\left(\frac{a+b}{2}\right)+f(b)\right)
\end{equation*}
for a step interval $[a,b]$. After the integration, the array shape becomes $s_{freq,sin}\in\mathbb{R}^{B\times f_{freq}T}$ where $B=6$ is the number of bands.

As our final step, we flatten the array for each channel to $s_{freq,sin,flat}\in\mathbb{R}^{Bf_{freq}T}$ and stack the $N$ channels together, resulting in shape $s_{freq}\in\mathbb{R}^{N\times Bf_{freq}T}$.

\paragraph{Slicing.}We select fixed window size $T_w=300$s. In this case, a microstates sample will have shape $c_w\in S^{f_{res}T_w}=S^{30000}$, and the raw EEG data will have shape $s_{time,w}\in\mathbb{R}^{N\times f_{res}T_w}=\mathbb{R}^{6\times 30000}$. The frequency-domain representation will have shape $s_{freq,w}\in\mathbb{R}^{N\times Bf_{freq}T_w}=\mathbb{R}^{6\times 1800}$.

\subsubsection{Labels.}
The label frequency is $f_l=\frac{1}{30}$Hz, and hence the label sequence will be of shape $l_w\in L^{f_lT_w}=L^{10}$ where $L$ consists of the five sleep stages. 

\subsection{Emotion Recognition}
\subsubsection{Dataset}
The dataset used is the \textbf{SEED} dataset \cite{duan2013differential,zheng2015investigating}. The SEED dataset consists of $15$ subjects whose EEG signals of $62$ channels are recorded when watching movie clips expressing different emotions which are categorized as positive, neutral and negative. There are a total number of $15$ trials, during which subjects view episodes with positive, neutral, negative, negative, nuetral, positive, negative, neutral, positive, positive, neutral, negative, neutral, positive, negative emotions. The dataset is filtered between $0$ and $75$Hz and downsampled to $200$Hz. In raw EEG samples have shape $s\in\mathbb{R}^{C\times f_sT}$ where $f_s=200$Hz and $C=62$. $T$ is the length of the movie clip which varies between trials.

\subsubsection{Preprocessing.}
\paragraph{Extracting target channels.}We extract the $6$ target channels as above for labeling. The resulting shape is $s_{ext}\in\mathbb{R}^{N\times f_sT}$ where $N=6$.

\paragraph{Constructing microstates and baseline.}We do not filter and resample the EEG signals since these are done initially. Applying the clustering model, we obtain the microstate sequence $c\in S^{f_sT}$ where $S$ is the set of $1000$ states. The raw signal has shape $s_{time}\in\mathbb{R}^{N\times f_sT}$, and the frequency-domain signal has shape $s_{freq}\in\mathbb{R}^{N\times Bf_{freq}T}$ where $f_{freq}=1$Hz and $B=6$ is the number of bands.

\paragraph{Windowing.}Since the movie clips are not of the same length, we set $T_w=265$s which is the duration of the longest video, and pad the signals that are shorter. For microstates, a new token is introduced for padding, whereas for the other two representations, we pad zeros. Now the microstate sequence has length $f_sT_w=53000$, the raw EEG signals have shape $s_{time,w}\in\mathbb{R}^{N\times f_sT_w}=\mathbb{R}^{6\times53000}$ and the frequency-domain signals have shape $s_{freq,w}\in\mathbb{R}^{6\times1590}$.

\subsubsection{Labels.}
\begin{figure}[ht]
    \centering
    \includegraphics[width=0.45\linewidth]{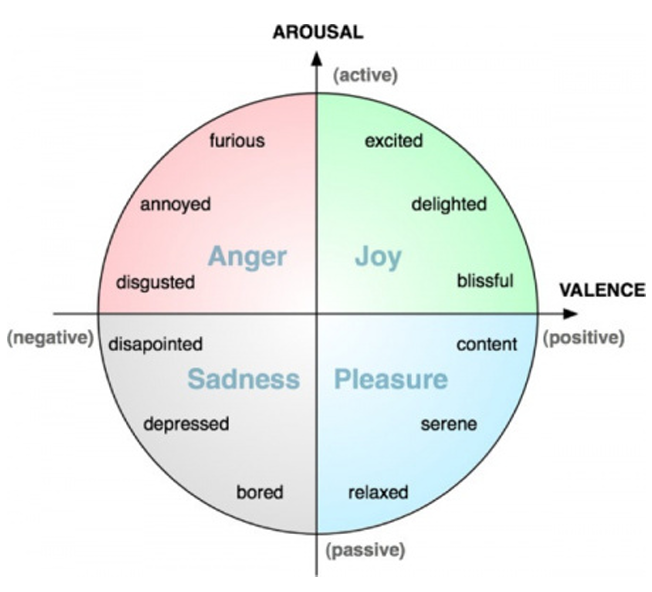}
    \caption{The Valence-Arousal Space}
    \label{fig:PAD}
\end{figure}

As mentioned in \cite{phdthesis}, human emotions can be characterized in the valence-arousal space as in \hyperref[fig:PAD]{Figure 7}. Valence and arousal are two dominant factors categorizing human feelings. Since the SEED dataset only features the valence aspect, the prediction of emotions is focused on the valence component, with labels $L$ defined as positive, neutral and negative. Each segmented window corresponds to a single emotion label.

\subsection{Motor Imagery Classification}
\subsubsection{Dataset}
We use the \textbf{Motor Movement/Imagery} dataset \cite{goldberger2000physionet,1300799}. The dataset consists of $109$ subjects undergoing $14$ trials. The $14$ trials includes two rest sessions and four tasks. The four tasks are:
\begin{itemize}
\item Task $1$: Open and close the left or right fist.
\item Task $2$: Imagine opening and closing the left or right fist.
\item Task $3$: Open and close both fists or both feet.
\item Task $4$: Imagine opening and closing both fists or feet.
\end{itemize}

Every subject went through two rest sessions and three rounds of successive tasks in the order above. The labels are given during movement roughly every four seconds. There are in total three labels. $T_0$ corresponds to rest, $T_1$ corresponds to the onset of moving or imagining moving the left or both fists, and $T_2$ corresponds to the onset of moving or imagining moving the right fist or both feet. The samples contain $64$ channels at $f_s=160$Hz. In this case, the raw EEG signals have shape $s\in\mathbb{R}^{C\times f_sT}$ where $C=64,f_s=160$Hz and $T$ is the duration of each trial.

\subsubsection{Preprocessing}
\paragraph{Extracting target channels.}Again, the six target channels are extracted and the resulting shape is $s_{ext}\in\mathbb{R}^{N\times f_sT},N=6$.

\paragraph{Constructing microstates and baseline.}We directly apply the clustering model on the raw EEG and obtain the microstate sequence $c\in S^{f_sT}$. The raw EEG signals have shape $s_{time}\in\mathbb{R}^{N\times f_sT}$ and the frequency-domain signals have shape $s_{freq}\in\mathbb{R}^{N\times Bf_{freq}T}$ where $f_{freq}=1$Hz.

\paragraph{Slicing.}Since each label lasts for roughly $4$s. We set $T_w=4$s. And thus the microstate sequence has length $640$, the raw EEG signal has shape $s_{time,w}\in\mathbb{R}^{6\times640}$ where as the frequency-domain signal has shape $s_{freq,w}\in\mathbb{R}^{6\times24}$.

\subsubsection{Labels.}
We let $L$ consists of four labels: left hand, right hand, both hands, both feet. Left hand corresponds to the label $T_1$ in trials $3,4,7,8,11,12$, right hand corresponds to the label $T_2$ in trials $3,4,7,8,11,12$, both hands corresponds to the label $T_1$ in trials $5,6,9,10,13,14$ and both feet corresponds to the label $T_2$ in trials $5,6,9,10,13,14$. Each sample corresponds to a single movement label.

\newpage
\section{Model Architecture and Training}
This section shows the detailed model structures adopted in this work.

\subsection{CNN+LSTM \cite{Shao2022AHD}}
\subsubsection{Model Details}
\paragraph{Overview of model structure.}The following shows the model structure. The three models have parameters $707$K, $692$K and $687$K respectively. Models are shown in \hyperref[table:cnn_lstm_raw_eeg]{Table $3$}, \hyperref[table:cnn_lstm_eeg]{Table $4$} and \hyperref[table:cnn_lstm_microstate]{Table $5$}.
\begin{table*}[ht!]
\vspace{0.1in}
\caption{CNN+LSTM for Raw EEG}
\vspace{0.1in}
\begin{center}
\begin{small}
\begin{sc}
\begin{tabular}{ccc}
\toprule
layer & output & configuration \\
\midrule
$-$ & $(6,30000)$ & $-$ \\
Conv1d & $(1024,30000)$ & input channels $6$, output channels $1024$, kernel size $5$ padding $2$ \\
BatchNorm1d & $(1024,30000)$ & $1024$ \\
Conv1d & $(128,10000)$ & input channels $1024$, output channels $128$, kernel size $3$ stride $3$ \\
MaxPool1d & $(128,5000)$ & kernel size $2$, stride $2$ \\
Dropout & $(128,5000)$ & $p=0.25$ \\
Conv1d & $(64,1000)$ & input channels $128$, output channels $64$, kernel size $5$ stride $5$ \\
MaxPool1d & $(64,500)$ & kernel size $2$, stride $2$ \\
Dropout & $(64,500)$ & $p=0.25$ \\
Conv1d & $(32,500)$ & input channels $64$, output channels $32$, kernel size $3$ padding $1$ \\
MaxPool1d & $(32,250)$ & kernel size $2$, stride $2$ \\
Dropout & $(32,250)$ & $p=0.25$ \\
GRU & $(64,250)$ & input size $32$, hidden size $64$, $2$ layers \\
Dropout & $(64,250)$ & $p=0.25$ \\
GRU & $(128,250)$ & input size $64$, hidden size $128$, $2$ layers \\
Dropout & $(128,250)$ & $p=0.25$ \\
reshape & $(10,3200)$ & $-$ \\
Linear & $(10,5)$ & input features $3200$, output features $5$ \\
\bottomrule
\end{tabular}
\end{sc}
\end{small}
\end{center}
\label{table:cnn_lstm_raw_eeg}
\end{table*}

\begin{table*}[ht!]
\vspace{0.1in}
\caption{CNN+LSTM for Frequency-Domain}
\vspace{0.1in}
\begin{center}
\begin{small}
\begin{sc}
\begin{tabular}{ccc}
\toprule
layer & output & configuration \\
\midrule
$-$ & $(6,1800)$ & $-$ \\
Conv1d & $(1024,1800)$ & input channels $6$, output channels $1024$, kernel size $5$ padding $2$ \\
BatchNorm1d & $(1024,1800)$ & $1024$ \\
Conv1d & $(128,600)$ & input channels $1024$, output channels $128$, kernel size $3$ stride $3$ \\
MaxPool1d & $(128,300)$ & kernel size $2$, stride $2$ \\
Dropout & $(128,300)$ & $p=0.25$ \\
Conv1d & $(64,60)$ & input channels $128$, output channels $64$, kernel size $5$ stride $5$ \\
MaxPool1d & $(64,30)$ & kernel size $2$, stride $2$ \\
Dropout & $(64,30)$ & $p=0.25$ \\
Conv1d & $(32,30)$ & input channels $64$, output channels $32$, kernel size $3$ padding $1$ \\
MaxPool1d & $(32,15)$ & kernel size $2$, stride $2$ \\
Dropout & $(32,15)$ & $p=0.25$ \\
GRU & $(64,15)$ & input size $32$, hidden size $64$, $2$ layers \\
Dropout & $(64,15)$ & $p=0.25$ \\
GRU & $(128,15)$ & input size $64$, hidden size $128$, $2$ layers \\
Dropout & $(128,15)$ & $p=0.25$ \\
reshape & $(10,192)$ & $-$ \\
Linear & $(10,5)$ & input features $192$, output features $5$ \\
\bottomrule
\end{tabular}
\end{sc}
\end{small}
\end{center}
\label{table:cnn_lstm_eeg}
\end{table*}

\begin{table*}[ht!]
\vspace{0.1in}
\caption{CNN+LSTM for Microstates}
\vspace{0.1in}
\begin{center}
\begin{small}
\begin{sc}
\begin{tabular}{ccc}
\toprule
layer & output & configuration \\
\midrule
$-$ & $(30000,)$ & $-$ \\
Embedding & $(30000,512)$ & number of embeddings $1000$, dimension $512$ \\
transpose & $(512,30000)$ & $-$ \\
BatchNorm1d & $(512,30000)$ & $512$ \\
Conv1d & $(64,10000)$ & input channels $512$, output channels $64$, kernel size $3$ stride $3$ \\
MaxPool1d & $(64,5000)$ & kernel size $2$, stride $2$ \\
Dropout & $(64,5000)$ & $p=0.25$ \\
Conv1d & $(32,1000)$ & input channels $64$, output channels $32$, kernel size $5$ stride $5$ \\
MaxPool1d & $(32,500)$ & kernel size $2$, stride $2$ \\
Dropout & $(32,500)$ & $p=0.25$ \\
Conv1d & $(16,500)$ & input channels $32$, output channels $16$, kernel size $3$ padding $1$ \\
MaxPool1d & $(16,250)$ & kernel size $2$, stride $2$ \\
Dropout & $(16,250)$ & $p=0.25$ \\
GRU & $(32,250)$ & input size $16$, hidden size $32$, $2$ layers \\
Dropout & $(32,250)$ & $p=0.25$ \\
GRU & $(64,250)$ & input size $32$, hidden size $64$, $2$ layers \\
Dropout & $(64,250)$ & $p=0.25$ \\
reshape & $(10,1600)$ & $-$ \\
Linear & $(10,5)$ & input features $1600$, output features $5$ \\
\bottomrule
\end{tabular}
\end{sc}
\end{small}
\end{center}
\label{table:cnn_lstm_microstate}
\end{table*}

\paragraph{CNN and GRU.}The convolution layers are employed to extract the spatial information across channels, and the gated recurrent units (GRUs) are used to extract temporal information.

GRU is a simplified version of long short term memory (LSTM) \cite{chung2014empiricalevaluationgatedrecurrent}. It consists of two gates---the update gate $z$ and the reset gate $r$.

At each time step $t$, the activation of the update gate $\boldsymbol{z}_t$ is computed as
\begin{equation*}
    \boldsymbol{z}_t=\sigma(W_z\boldsymbol{x}_t+U_z\boldsymbol{h}_{t-1})
\end{equation*}
where $h_{t-1}$ is the activation of the GRU at time step $t-1$ and $\sigma$ denotes the element-wise sigmoid function. Similarly, the activation $\boldsymbol{r}_t$ of the reset gate is computed as
\begin{equation*}
    \boldsymbol{r}_t=\sigma(W_r\boldsymbol{x}_t+U_r\boldsymbol{h}_{t-1})
\end{equation*}

Next, the candidate activate $\tilde{h}_t$ is computed as
\begin{equation*}
    \boldsymbol{\tilde{h}}_t=\tanh(W\boldsymbol{x}_t+\boldsymbol{r}_t^TU\boldsymbol{h}_{t-1})
\end{equation*}
The activation $h_t$ at time step $t$ is computed as
\begin{equation*}
    \boldsymbol{h}_t=(\boldsymbol{1}^T-\boldsymbol{z}_t^T)\boldsymbol{h}_{t-1}+\boldsymbol{z}_t^T\boldsymbol{\tilde{h}}_{t-1}
\end{equation*}

Using this mechanism, the model can selectively consider input at different time steps.

\paragraph{Embedding layer.} To adapt the model simultaneously to continuous and discrete EEG representations, we use different layers for microstates and conventional representations. For microstates, an embedding layer is adopted to convert discrete microstates into high-dimensional vectors, and for continuous signals, we use a convolution layer, which functions similarly by mapping the input into a high-dimensional latent space. The dimensions are chosen appropriately to guarantee that the model parameters are roughly the same.

\subsubsection{Training Configuration}
This section lists the training configurations of the above $3$ models in \hyperref[table:cnn_lstm_config]{Table $6$}. The models are trained on an NVIDIA-H$20$ GPU. The parameters in each case is optimized for performance and memory utilization.

\begin{table*}[t]
\vspace{0.1in}
\caption{Training Configuration for CNN+LSTM}
\vspace{0.1in}
\begin{center}
\begin{small}
\begin{sc}
\begin{tabular}{cccccc}
\toprule
parameter&batch&optimizer&learning rate&split (train:val:test)&early stop\\
\midrule
Raw EEG&$64$&Adam&$10^{-4}$&$7:1:2$&patience $20$ on Kappa\\
Frequency-Domain&$256$&Adam&$10^{-4}$&$7:1:2$&patience $20$ on Kappa\\
Microstates&$512$&Adam&$10^{-4}$&$7:1:2$&patience $20$ on Kappa\\
\bottomrule
\end{tabular}
\end{sc}
\end{small}
\end{center}
\label{table:cnn_lstm_config}
\end{table*}
\subsection{Sleep Transformer \cite{9697331}}
\subsubsection{Model Details}
\paragraph{Overview of model structure.}The following shows the model structure of Sleep Transformer. Parameters are $3.2$M, $3.2$M and $3.4$M, respectively. Model structures are shown in \hyperref[table:sleep_trans_raw_eeg]{Table $7$}, \hyperref[table:sleep_trans_eeg]{Table $8$} and \hyperref[table:sleep_trans_microstate]{Table $9$}.
\begin{table*}[ht!]
\caption{Sleep Transformer for Raw EEG}
\vspace{0.1in}
\begin{center}
\begin{small}
\begin{sc}
\begin{tabular}{ccc}
\toprule
layer & output & configuration \\
\midrule
$-$&$(b,6,30000)$&$-$ \\
Conv1d&$(b,6,6000)$&input channels $6$, output channels $6$, kernel size $5$ stride $5$ \\
Conv1d&$(b,6,3000)$&input channels $6$, output channels $6$, kernel size $2$ stride $2$ \\
transpose&$(b,3000,6)$&$-$ \\
reshape&$(10b,300,6)$&$-$ \\
RoFormer&$(10b,300,256)$&hidden size $256$, $2$ hidden layers, $4$ heads, intermediate size $1024$ \\
slice and reshape&$(b,10,256)$&retrieve only the first along the second dimension and reshape \\
RoFormer&$(b,10,256)$&hidden size $256$, $2$ hidden layers, $4$ heads, intermediate size $1024$ \\
Linear&$(b,10,5)$&input features $256$, output features $5$ \\
\bottomrule
\end{tabular}
\end{sc}
\end{small}
\end{center}
\label{table:sleep_trans_raw_eeg}
\end{table*}

\begin{table*}[ht!]
\caption{Sleep Transformer for Frequency-Domain}
\vspace{0.1in}
\begin{center}
\begin{small}
\begin{sc}
\begin{tabular}{ccc}
\toprule
layer & output & configuration \\
\midrule
$-$&$(b,6,1800)$&$-$ \\
transpose&$(b,1800,6)$&$-$ \\
reshape&$(10b,180,6)$&$-$ \\
RoFormer&$(10b,180,256)$&hidden size $256$, $2$ hidden layers, $4$ heads, intermediate size $1024$ \\
slice and reshape&$(b,10,256)$&retrieve only the first along the second dimension and reshape \\
RoFormer&$(b,10,256)$&hidden size $256$, $2$ hidden layers, $4$ heads, intermediate size $1024$ \\
Linear&$(b,10,5)$&input features $256$, output features $5$ \\
\bottomrule
\end{tabular}
\end{sc}
\end{small}
\end{center}
\label{table:sleep_trans_eeg}
\end{table*}

\begin{table*}[ht!]
\caption{Sleep Transformer for Microstates}
\vspace{0.1in}
\begin{center}
\begin{small}
\begin{sc}
\begin{tabular}{ccc}
\toprule
layer & output & configuration \\
\midrule
$-$&$(b,30000)$&$-$ \\
Embedding&$(b,30000,128)$&number of embeddings $1002$, dimension $128$ \\
transpose&$(b,128,30000)$&$-$ \\
Conv1d&$(b,128,6000)$&input channels $128$, output channels $128$, kernel size $5$ stride $5$ \\
Conv1d&$(b,128,3000)$&input channels $128$, output channels $128$, kernel size $2$ stride $2$ \\
transpose&$(b,3000,128)$&$-$ \\
reshape&$(10b,300,128)$&$-$ \\
RoFormer&$(10b,300,256)$&hidden size $256$, $2$ hidden layers, $4$ heads, intermediate size $1024$ \\
slice and reshape&$(b,10,256)$&retrieve only the first along the second dimension and reshape \\
RoFormer&$(b,10,256)$&hidden size $256$, $2$ hidden layers, $4$ heads, intermediate size $1024$ \\
Linear&$(b,10,5)$&input features $256$, output features $5$ \\
\bottomrule
\end{tabular}
\end{sc}
\end{small}
\end{center}
\label{table:sleep_trans_microstate}
\end{table*}

\paragraph{RoFormer.}The main part of the model uses an attention-based mechanism to extract temporal features. RoFormer is proposed in \cite{su2023roformerenhancedtransformerrotary}, which utilizes a novel positional embedding.

Generally speaking, the attention mechanism needs a key $\boldsymbol{k}_i$, query $\boldsymbol{q}_i$ and value $\boldsymbol{v}_i$ for each input position $i$. We can write them as
\begin{equation*}
    \begin{aligned}
        \boldsymbol{q}_i&=f_q(\boldsymbol{x}_i,i)\\
        \boldsymbol{k}_i&=f_k(\boldsymbol{x}_i,i)\\
        \boldsymbol{v}_i&=f_v(\boldsymbol{x}_i,i)\\
    \end{aligned}
\end{equation*}
where $\boldsymbol{x}_i$ is the word vector at position $i$. The attention between position $m,n$ is calculated as
\begin{equation*}
    a_{m,n}=\frac{e^\frac{\boldsymbol{q}_m^T\boldsymbol{k}_n}{\sqrt{d}}}{\sum_{j=1}^Ne^\frac{\boldsymbol{q}_m^T\boldsymbol{k}_j}{\sqrt{d}}}
\end{equation*}
and since this value is calculated in parallel, we have to incorporate the positional information $i$ along with $\boldsymbol{x}_i$ into the queries and keys.

The main idea of RoFormer is to select a positional embedding such that
\begin{equation*}
    \boldsymbol{q}_m^T\boldsymbol{k}_n=g(\boldsymbol{x}_m,\boldsymbol{x}_n,n-m)
\end{equation*}
is a function that depends solely on the input word vector and the relative position between $m,n$.

To construct such a positional embedding, let the embedding dimension be $d$ which is an even number, then we construct the following matrix
\begin{equation*}
    \boldsymbol{R}^d_{\Theta,m}=\begin{pmatrix}\cos m\theta_1&-\sin m\theta_1&\ldots&0&0\\\sin m\theta_1&\cos m\theta_1&\ldots&0&0\\0&0&\ldots&0&0\\0&0&\ldots&0&0\\\vdots&\vdots&\ddots&\vdots&\vdots\\0&0&\ldots&\cos m\theta_\frac{d}{2}&-\sin m\theta_\frac{d}{2}\\0&0&\ldots&\sin m\theta_\frac{d}{2}&\cos m\theta_\frac{d}{2}\end{pmatrix}
\end{equation*}
and we have
\begin{equation*}
    \begin{aligned}
    \boldsymbol{q}_m&=\boldsymbol{R}^d_{\Theta,m}\boldsymbol{W}_q\boldsymbol{x}_m\\
    \boldsymbol{k}_n&=\boldsymbol{R}^d_{\Theta,n}\boldsymbol{W}_k\boldsymbol{x}_k\\
    \boldsymbol{q}_m^T\boldsymbol{k}_n&=\boldsymbol{x}_m^T\boldsymbol{W}_q^T\boldsymbol{R}^d_{\Theta,n-m}\boldsymbol{W}_k\boldsymbol{x}_k=g(\boldsymbol{x}_m,\boldsymbol{x}_n,n-m)
    \end{aligned}
\end{equation*}

\paragraph{Embedding layer.}An embedding layer is added before the microstates model. There are $2$ extra vectors for padding and classification token. A convolution layer is used instead for continuous signals.

\subsubsection{Training Configuration}
The models are trained on an NVIDIA-H$20$ GPU. Parameters are optimized for performance and memory utilization.

\begin{table*}[ht!]
\vspace{0.1in}
\caption{Training Configuration for Sleep Transformer}
\vspace{0.1in}
\begin{center}
\begin{small}
\begin{sc}
\begin{tabular}{cccccc}
\toprule
parameter&batch&optimizer&learning rate&split (train:val:test)&early stop\\
\midrule
Raw EEG&$64$&Adam&$10^{-4}$&$7:1:2$&patience $20$ on Kappa\\
Frequency-Domain&$1000$&Adam&$10^{-4}$&$7:1:2$&patience $20$ on Kappa\\
Microstates&$512$&Adam&$10^{-4}$&$7:1:2$&patience $20$ on Kappa\\
\bottomrule
\end{tabular}
\end{sc}
\end{small}
\end{center}
\label{table:sleep_trans_config}
\end{table*}

\subsection{Sleep Net Zero \cite{li2024sleepnetzero}}
\subsubsection{Model Details}
\paragraph{Overview of model structure.}The following shows the model structure of Sleep Net Zero. Parameters are $10.9$M, $3.2$M and $3.2$M. Model details are shown in \hyperref[table:sleep_net_zero_raw_eeg]{Table $12$}, \hyperref[table:sleep_net_zero_eeg]{Table $13$} and \hyperref[table:sleep_net_zero_microstate]{Table $14$}.

\begin{figure}[ht]
    \centering
    \includegraphics[width=0.48\linewidth]{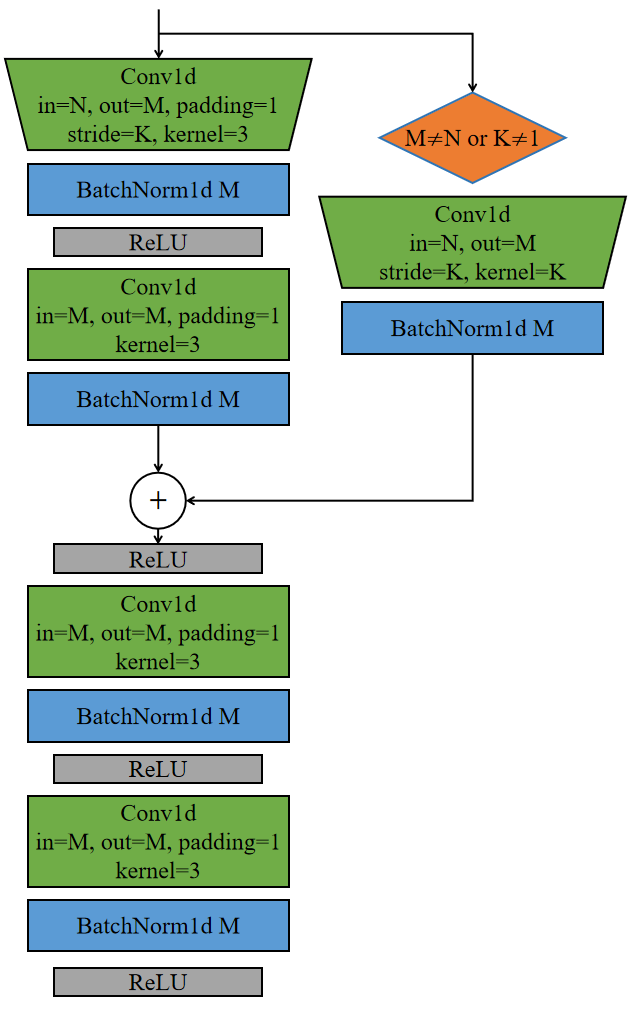}
    \caption{The Model Structure of ResBlocks with in=N, out=M and stride=K}
    \label{fig:wuji_res}
\end{figure}

\begin{table*}[ht!]
\caption{ResNetFeatureExtractor}
\vspace{0.1in}
\begin{center}
\begin{small}
\begin{sc}
\begin{tabular}{ccc}
\toprule
layer & output & configuration \\
\midrule
$-$&$(6,30000)$&$-$ \\
Conv1d&$(64,30000)$&input channels $6$, output channels $64$, kernel size $7$ padding $3$ \\
BatchNorm1d&$(64,30000)$&$64$ \\
ReLU&$(64,30000)$&$-$ \\
MaxPool1d&$(64,15000)$&kernel size $3$ stride $2$ padding $1$ \\
ResBlocks&$(512,3000)$&ResBlocks(in=$64$,out=$512$,stride=$5$), see \hyperref[fig:wuji_res]{above} \\
ResBlocks&$(128,3000)$&ResBlocks(in=$512$,out=$128$,stride=$1$), see \hyperref[fig:wuji_res]{above} \\
ResBlocks&$(256,3000)$&ResBlocks(in=$128$,out=$256$,stride=$1$), see \hyperref[fig:wuji_res]{above} \\
ResBlocks&$(512,3000)$&ResBlocks(in=$256$,out=$512$,stride=$1$), see \hyperref[fig:wuji_res]{above} \\
\bottomrule
\end{tabular}
\end{sc}
\end{small}
\end{center}
\label{table:res_net_feature_extractor}
\end{table*}

\begin{table*}[ht!]
\vspace{0.1in}
\caption{Sleep Net Zero for Raw EEG}
\vspace{0.1in}
\begin{center}
\begin{small}
\begin{sc}
\begin{tabular}{ccc}
\toprule
layer & output & configuration \\
\midrule
$-$&$(6,30000)$&$-$ \\
ResNetFeatureExtractor&$(512,3000)$&see \hyperref[table:res_net_feature_extractor]{above} \\
transpose&$(3000,512)$&$-$ \\
RoFormer&$(3000,512)$&hidden size $512$, $2$ hidden layers, $2$ heads, intermediate size $1024$ \\
Linear&$(3000,5)$&input features $512$, output features $5$ \\
reshape and mean&$(10,5)$&compute the mean of every consecutive $300$ scores \\
\bottomrule
\end{tabular}
\end{sc}
\end{small}
\end{center}
\label{table:sleep_net_zero_raw_eeg}
\end{table*}

\begin{table*}[ht!]
\caption{Sleep Net Zero for Frequency-Domain}
\vspace{0.1in}
\begin{center}
\begin{small}
\begin{sc}
\begin{tabular}{ccc}
\toprule
layer & output & configuration \\
\midrule
$-$&$(6,1800)$&$-$ \\
Conv1d&$(640,1800)$&input channels $6$, output channels $640$, kernel size $5$ padding $2$ \\
Conv1d&$(640,360)$&input channels $640$, output channels $640$, kernel size $5$ padding $5$ \\
Conv1d&$(320,360)$&input channels $640$, output channels $320$, kernel size $3$ padding $1$ \\
Conv1d&$(320,180)$&input channels $320$, output channels $320$, kernel size $2$ padding $2$ \\
Conv1d&$(160,180)$&input channels $320$, output channels $160$, kernel size $3$ padding $1$ \\
transpose&$(180,160)$&$-$ \\
RoFormer&$(180,160)$&hidden size $160$, $2$ hidden layers, $2$ heads, intermediate size $1024$ \\
Linear&$(180,5)$&input features $160$, output features $5$ \\
reshape and mean&$(10,5)$&compute the mean of every consecutive $18$ scores \\
\bottomrule
\end{tabular}
\end{sc}
\end{small}
\end{center}
\label{table:sleep_net_zero_eeg}
\end{table*}

\begin{table*}[ht!]
\caption{Sleep Net Zero for Microstates}
\vspace{0.1in}
\begin{center}
\begin{small}
\begin{sc}
\begin{tabular}{ccc}
\toprule
layer & output & configuration \\
\midrule
$-$&$(30000,)$&$-$ \\
Embedding&$(30000,512)$&number of embeddings $1002$, dimension $512$ \\
transpose&$(512,30000)$&$-$ \\
Conv1d&$(512,6000)$&input channels $512$, output channels $512$, kernel size $5$ padding $5$ \\
Conv1d&$(256,6000)$&input channels $512$, output channels $256$, kernel size $3$ padding $1$ \\
Conv1d&$(256,3000)$&input channels $256$, output channels $256$, kernel size $2$ padding $2$ \\
Conv1d&$(128,3000)$&input channels $256$, output channels $128$, kernel size $3$ padding $1$ \\
transpose&$(3000,128)$&$-$ \\
RoFormer&$(3000,128)$&hidden size $128$, $2$ hidden layers, $2$ heads, intermediate size $1024$ \\
Linear&$(3000,5)$&input features $128$, output features $5$ \\
reshape and mean&$(10,5)$&compute the mean of every consecutive $300$ scores \\
\bottomrule
\end{tabular}
\end{sc}
\end{small}
\end{center}
\label{table:sleep_net_zero_microstate}
\end{table*}

\paragraph{Embedding layer.}We adopt an embedding layer with $1002$ tokens for microstates. For raw EEG signals, since its input size is significantly larger than the other two representations, we increase the embedding dimension for better performance.

\subsubsection{Training Configuration}
The models are trained on an NVIDIA-H$20$ GPU. Parameters in each case are optimized for performance and memory utilization.

\begin{table*}[ht!]
\caption{Training Configuration for Sleep Net Zero}
\vspace{0.1in}
\begin{center}
\begin{small}
\begin{sc}
\begin{tabular}{cccccc}
\toprule
parameter&batch&optimizer&learning rate&split (train:val:test)&early stop\\
\midrule
Raw EEG&$96$&Adam&$10^{-4}$&$7:1:2$&patience $20$ on Kappa\\
Frequency-Domain&$512$&Adam&$10^{-4}$&$7:1:2$&patience $20$ on Kappa\\
Microstates&$128$&Adam&$10^{-4}$&$7:1:2$&patience $20$ on Kappa\\
\bottomrule
\end{tabular}
\end{sc}
\end{small}
\end{center}
\label{table:sleep_net_zero_config}
\end{table*}

\subsection{CNN-Based Model for Emotion Recognition \cite{Cheng2020SubjectAwareCL}}
Apart from sleep staging, we show the model used for emotion recognition.
\subsubsection{Model Details}
\paragraph{Overview of model structure.}The following shows the model structure of the CNN-based model. Parameters are $19.1$M, $19.1$M and $20.1$M. Model details are shown in \hyperref[table:cnn_raw_eeg]{Table $16$}, \hyperref[table:cnn_eeg]{Table $17$} and \hyperref[table:cnn_microstate]{Table $18$}.

\begin{table*}[ht!]
\vspace{0.1in}
\caption{CNN for Raw EEG}
\vspace{0.1in}
\begin{center}
\begin{small}
\begin{sc}
\begin{tabular}{ccc}
\toprule
layer & output & configuration \\
\midrule
$-$ & $(6,53000)$ & $-$ \\
Conv1d & $(1024,53000)$ & input channels $6$, output channels $1024$, kernel size $1$ \\
Conv1d and ReLU & $(256,10600)$ & input channels $1024$, output channels $256$, kernel size $5$ stride $5$ \\
Conv1d and ReLU & $(128,2120)$ & input channels $256$, output channels $128$, kernel size $5$ stride $5$ \\
Conv1d and ReLU & $(128,1060)$ & input channels $128$, output channels $128$, kernel size $2$ stride $2$ \\
MaxPool1d & $(128,530)$&kernel size $2$ stride $2$\\
Dropout & $(128,530)$&$p=0.1$ \\
Conv1d and ReLU & $(128,530)$ & input channels $128$, output channels $128$, kernel size $3$ padding $1$ \\
Conv1d and ReLU & $(128,530)$ & input channels $128$, output channels $128$, kernel size $3$ padding $1$ \\
MaxPool1d & $(128,265)$&kernel size $2$ stride $2$ \\
Dropout & $(128,265)$ & $p=0.1$ \\
flatten & $(33920,)$ & $-$ \\
Linear & $(512,)$ & input features $33920$, output features $512$ \\
Dropout& $(512,)$ & $p=0.1$ \\
Linear & $(256,)$ & input features $512$, output features $256$ \\
Dropout& $(256,)$ & $p=0.1$ \\
Linear & $(64,)$ & input features $256$, output features $64$ \\
Dropout& $(64,)$ & $p=0.1$ \\
Linear & $(3,)$ & input features $64$, output features $3$ \\
\bottomrule
\end{tabular}
\end{sc}
\end{small}
\end{center}
\label{table:cnn_raw_eeg}
\end{table*}

\begin{table*}[ht!]
\vspace{0.1in}
\caption{CNN for Frequency-Domain}
\vspace{0.1in}
\begin{center}
\begin{small}
\begin{sc}
\begin{tabular}{ccc}
\toprule
layer & output & configuration \\
\midrule
$-$ & $(6,1590)$ & $-$ \\
reshape & $(6,265,6)$ & let the last dimension be the frequency bands \\
Conv2d & $(1024,265,6)$ & input channels $6$, output channels $1024$, kernel size $(1,1)$ \\
Conv2d \& ReLU & $(256,265,6)$ & input channels $1024$, output channels $256$, kernel size $(1,5)$ padding $(0,2)$ \\
Conv2d \& ReLU & $(128,265,6)$ & input channels $256$, output channels $128$, kernel size $(1,5)$ stride $(0,2)$ \\
Conv2d \& ReLU & $(128,265,6)$ & input channels $128$, output channels $128$, kernel size $(1,3)$ stride $(0,1)$ \\
MaxPool2d & $(128,265,3)$&kernel size $(1,2)$ stride $(1,2)$\\
Dropout & $(128,265,3)$&$p=0.1$ \\
Conv2d \& ReLU & $(128,265,3)$ & input channels $128$, output channels $128$, kernel size $(1,3)$ padding $(0,1)$ \\
Conv2d \& ReLU & $(128,265,3)$ & input channels $128$, output channels $128$, kernel size $(1,3)$ padding $(0,1)$ \\
MaxPool2d & $(128,265,1)$&kernel size $(1,2)$ stride $(1,2)$ \\
Dropout & $(128,265,1)$ & $p=0.1$ \\
flatten & $(33920,)$ & $-$ \\
Linear & $(512,)$ & input features $33920$, output features $512$ \\
Dropout& $(512,)$ & $p=0.1$ \\
Linear & $(256,)$ & input features $512$, output features $256$ \\
Dropout& $(256,)$ & $p=0.1$ \\
Linear & $(64,)$ & input features $256$, output features $64$ \\
Dropout& $(64,)$ & $p=0.1$ \\
Linear & $(3,)$ & input features $64$, output features $3$ \\
\bottomrule
\end{tabular}
\end{sc}
\end{small}
\end{center}
\label{table:cnn_eeg}
\end{table*}

\begin{table*}[ht!]
\vspace{0.1in}
\caption{CNN for Microstates}
\vspace{0.1in}
\begin{center}
\begin{small}
\begin{sc}
\begin{tabular}{ccc}
\toprule
layer & output & configuration \\
\midrule
$-$ & $(6,53000)$ & $-$ \\
Embedding & $(1024,53000)$ & number of embeddings $1001$, dimension $1024$ \\
Conv1d and ReLU & $(256,10600)$ & input channels $1024$, output channels $256$, kernel size $5$ stride $5$ \\
Conv1d and ReLU & $(128,2120)$ & input channels $256$, output channels $128$, kernel size $5$ stride $5$ \\
Conv1d and ReLU & $(128,1060)$ & input channels $128$, output channels $128$, kernel size $2$ stride $2$ \\
MaxPool1d & $(128,530)$&kernel size $2$ stride $2$\\
Dropout & $(128,530)$&$p=0.1$ \\
Conv1d and ReLU & $(128,530)$ & input channels $128$, output channels $128$, kernel size $3$ padding $1$ \\
Conv1d and ReLU & $(128,530)$ & input channels $128$, output channels $128$, kernel size $3$ padding $1$ \\
MaxPool1d & $(128,265)$&kernel size $2$ stride $2$ \\
Dropout & $(128,265)$ & $p=0.1$ \\
flatten & $(33920,)$ & $-$ \\
Linear & $(512,)$ & input features $33920$, output features $512$ \\
Dropout& $(512,)$ & $p=0.1$ \\
Linear & $(256,)$ & input features $512$, output features $256$ \\
Dropout& $(256,)$ & $p=0.1$ \\
Linear & $(64,)$ & input features $256$, output features $64$ \\
Dropout& $(64,)$ & $p=0.1$ \\
Linear & $(3,)$ & input features $64$, output features $3$ \\
\bottomrule
\end{tabular}
\end{sc}
\end{small}
\end{center}
\label{table:cnn_microstate}
\end{table*}

\paragraph{Embedding layer.}For microstates, an embedding layer with vocabulary $1001$ and dimension $1024$ is employed. The extra token is for padding. Convolution layers are used in the place of embedding for the other two representations.
\subsubsection{Training Configuration}
This section lists the training configurations of the above three models. The models are trained on an NVIDIA-H$20$ GPU. The parameters in each case are optimized for performance and memory utilization.

\begin{table*}[ht!]
\caption{Training Configuration for CNN}
\vspace{0.1in}
\begin{center}
\begin{small}
\begin{sc}
\begin{tabular}{cccccc}
\toprule
parameter&batch&optimizer&learning rate&split (train:val:test)&early stop\\
\midrule
Raw EEG&$128$&Adam&$5\times10^{-4}$&$7:1:2$&patience $100$ on Kappa\\
Frequency-Domain&$128$&Adam&$5\times10^{-4}$&$7:1:2$&patience $100$ on Kappa\\
Microstates&$128$&Adam&$10^{-4}$&$7:1:2$&patience $100$ on Kappa\\
\bottomrule
\end{tabular}
\end{sc}
\end{small}
\end{center}
\label{table:cnn_config}
\end{table*}

\subsection{ResNet Model for Motor Imagery Classification}
Finally, we list our model for motor imagery classification.
\subsubsection{Model Details}
\paragraph{Overview of model structure.}The following shows the structure of ResNet model. Parameters are $20.3$M, $21.5$M and $21.4$M. Model details are shown in \hyperref[table:res_net_raw_eeg]{Table $20$}, \hyperref[table:res_net_eeg]{Table $21$} and \hyperref[table:res_net_microstate]{Table $22$}.

\begin{table*}[ht!]
\vspace{0.1in}
\caption{ResNet for Raw EEG}
\vspace{0.1in}
\begin{center}
\begin{small}
\begin{sc}
\begin{tabular}{ccc}
\toprule
layer & output & configuration \\
\midrule
$-$ & $(6,640)$ & $-$ \\
Conv1d&$(1024,640)$&input channels $6$, output channels $1024$, kernel size $3$ padding $1$ \\
Encoder&$(128,640)$&see \hyperref[table:encoder]{below} \\
flatten&$(81920,)$&$-$ \\
Classifier&$(4,)$&see \hyperref[table:classifier]{below} \\
\bottomrule
\end{tabular}
\end{sc}
\end{small}
\end{center}
\label{table:res_net_raw_eeg}
\end{table*}

\begin{table*}[ht!]
\vspace{0.1in}
\caption{ResNet for Frequency-Domain}
\vspace{0.1in}
\begin{center}
\begin{small}
\begin{sc}
\begin{tabular}{ccc}
\toprule
layer & output & configuration \\
\midrule
$-$ & $(6,24)$ & $-$ \\
Conv1d&$(1024,24)$&input channels $6$, output channels $1024$, kernel size $3$, padding $1$ \\
Encoder $2$&$(256,24)$&see \hyperref[table:encoder_2]{below} \\
flatten&$(6144,)$&$-$ \\
Classifier $2$&$(4,)$&see \hyperref[table:classifier_2]{below} \\
\bottomrule
\end{tabular}
\end{sc}
\end{small}
\end{center}
\label{table:res_net_eeg}
\end{table*}

\begin{table*}[ht!]
\vspace{0.1in}
\caption{ResNet for Microstates}
\vspace{0.1in}
\begin{center}
\begin{small}
\begin{sc}
\begin{tabular}{ccc}
\toprule
layer & output & configuration \\
\midrule
$-$ & $(6,640)$ & $-$ \\
Conv1d&$(1024,640)$&input channels $6$, output channels $1024$, kernel size $3$, padding $1$ \\
Encoder&$(128,640)$&see \hyperref[table:encoder]{below} \\
flatten&$(81920,)$&$-$ \\
Classifier&$(4,)$&see \hyperref[table:classifier]{below} \\
\bottomrule
\end{tabular}
\end{sc}
\end{small}
\end{center}
\label{table:res_net_microstate}
\end{table*}

\begin{table*}[ht!]
\caption{Encoder Architecture}
\vspace{0.1in}
\begin{center}
\begin{small}
\begin{sc}
\begin{tabular}{ccc}
\toprule
layer & output & configuration \\
\midrule
$-$ & $(1024,640)$ & $-$ \\
Conv1d&$(512,640)$&input channels $1024$, output channels $512$, kernel size $13$, padding $6$ \\
ResBlock1d&$(256,640)$&in $512$, out $256$, kernel $11$, see \hyperref[fig:resblock]{below} \\
ResBlock1d&$(128,640)$&in $256$, out $128$, kernel $9$, see \hyperref[fig:resblock]{below} \\
ResBlock1d&$(128,640)$&in $128$, out $128$, kernel $7$, see \hyperref[fig:resblock]{below} \\
ELU&$(128,640)$&$-$ \\
\bottomrule
\end{tabular}
\end{sc}
\end{small}
\end{center}
\label{table:encoder}
\end{table*}

\begin{table*}[ht!]
\caption{Encoder $2$ Architecture}
\vspace{0.1in}
\begin{center}
\begin{small}
\begin{sc}
\begin{tabular}{ccc}
\toprule
layer & output & configuration \\
\midrule
$-$ & $(1024,24)$ & $-$ \\
Conv1d&$(768,24)$&input channels $1024$, output channels $768$, kernel size $13$, padding $6$ \\
ResBlock1d&$(512,24)$&in $768$, out $512$, kernel $11$, see \hyperref[fig:resblock]{below} \\
ResBlock1d&$(256,24)$&in $512$, out $256$, kernel $9$, see \hyperref[fig:resblock]{below} \\
ResBlock1d&$(256,24)$&in $256$, out $256$, kernel $7$, see \hyperref[fig:resblock]{below} \\
ELU&$(256,24)$&$-$ \\
\bottomrule
\end{tabular}
\end{sc}
\end{small}
\end{center}
\label{table:encoder_2}
\end{table*}

\begin{figure}[ht]
    \centering
    \includegraphics[width=0.4\linewidth]{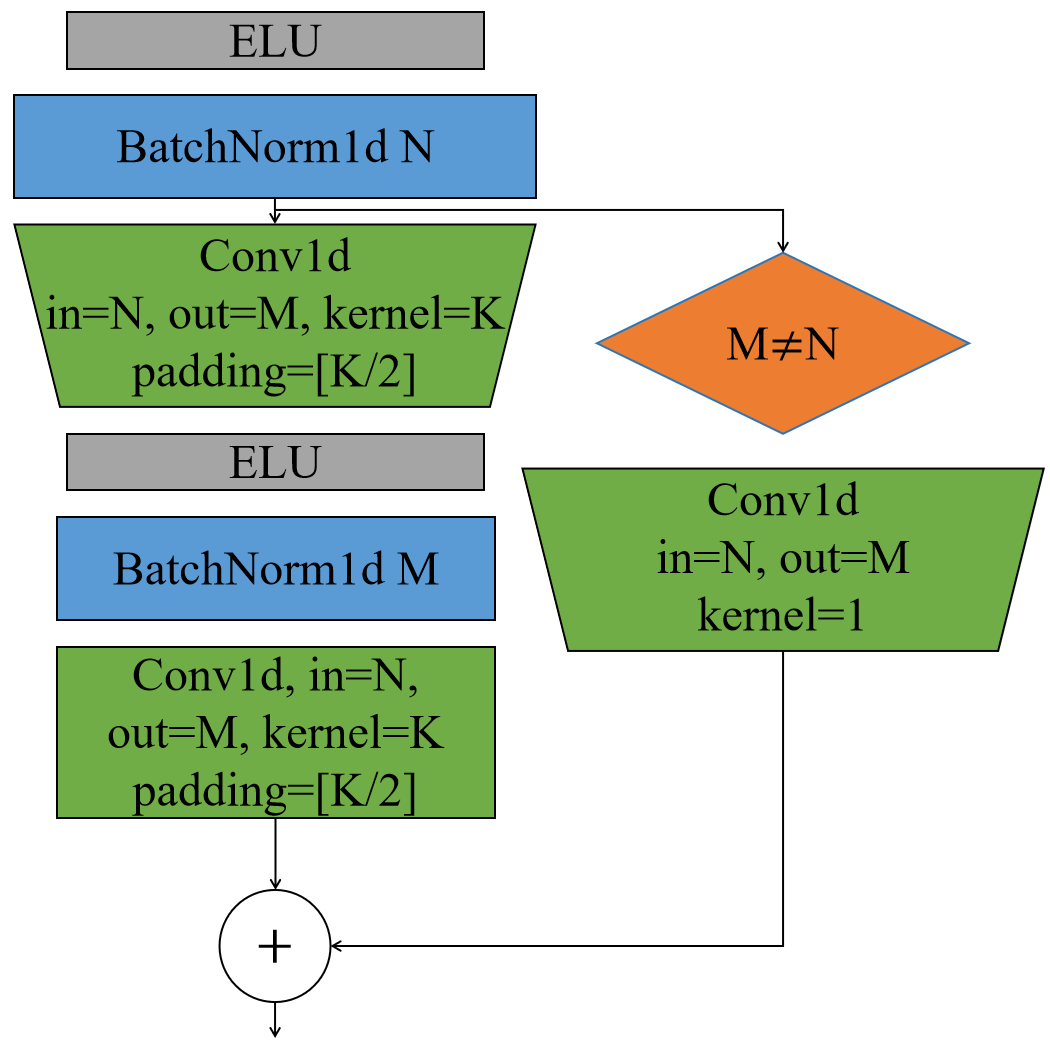}
    \caption{The Model Structure of ResBlock1d with in=N, out=M and kernel=K}
    \label{fig:resblock}
\end{figure}

\begin{table*}[ht!]
\caption{Classifier Architecture}
\vspace{0.1in}
\begin{center}
\begin{small}
\begin{sc}
\begin{tabular}{ccc}
\toprule
layer & output & configuration \\
\midrule
$-$ & $(81920,)$ & $-$ \\
Linear and ReLU & $(128,)$ & in features $81920$, out features $128$ \\
Linear and ReLU & $(128,)$ & in features $128$, out features $128$ \\
Linear and ReLU & $(64,)$ & in features $128$, out features $64$ \\
Linear & $(4,)$ & in features $64$, out features $4$ \\
\bottomrule
\end{tabular}
\end{sc}
\end{small}
\end{center}
\label{table:classifier}
\end{table*}

\begin{table*}[ht!]
\caption{Classifier $2$ Architecture}
\vspace{0.1in}
\begin{center}
\begin{small}
\begin{sc}
\begin{tabular}{ccc}
\toprule
layer & output & configuration \\
\midrule
$-$ & $(6144,)$ & $-$ \\
Linear and ReLU & $(128,)$ & in features $6144$, out features $128$ \\
Linear and ReLU & $(128,)$ & in features $128$, out features $128$ \\
Linear and ReLU & $(64,)$ & in features $128$, out features $64$ \\
Linear & $(4,)$ & in features $64$, out features $4$ \\
\bottomrule
\end{tabular}
\end{sc}
\end{small}
\end{center}
\label{table:classifier_2}
\end{table*}

\paragraph{ELU.}The ELU activation function is defined as
\begin{equation*}
    ELU(x)=\left\{
    \begin{aligned}
    x&&x>0\\
    \alpha(e^x-1)&&x\le0
    \end{aligned}
    \right.
\end{equation*}
\paragraph{Embedding layer.}For microstates, an embedding layer with vocabulary $1000$ and dimension $1024$ is employed. Convolution layers are used in the place of embedding for the other two representations.
\subsubsection{Training Configuration}
This section lists the training configurations of the above $3$ models. The models are trained on an NVIDIA-H$20$ GPU. Model parameters in each case are optimized for performance and memory utilization.

\begin{table*}[ht!]
\caption{Training Configuration for ResNet}
\vspace{0.1in}
\begin{center}
\begin{small}
\begin{sc}
\begin{tabular}{cccccc}
\toprule
parameter&batch&optimizer&learning rate&split (train:val:test)&early stop\\
\midrule
Raw EEG&$128$&Adam&$5\times10^{-4}$&$7:1:2$&patience $100$ on Kappa\\
Frequency-Domain&$128$&Adam&$5\times10^{-4}$&$7:1:2$&patience $100$ on Kappa\\
Microstates&$128$&Adam&$2\times10^{-6}$&$7:1:2$&patience $10$ on Kappa\\
\bottomrule
\end{tabular}
\end{sc}
\end{small}
\end{center}
\label{table:res_net_config}
\end{table*}

\newpage
\section{More Microstate Analysis}\label{appendix:C}
This section provides more analysis and visualization of microstates.
\subsection{Comparison Between Wake Stage and Rapid Eye Movement (REM) Stage}
Humans undergo vivid dreaming processes during REM stage \cite{Vaudano2019}. In turn, EEG signals in REM stage share the same characteristics with that during wakefulness. We analyze the $30$ most frequent-appearing microstates during W and REM stages across groups of $30$ subjects. The results are as follows:

\begin{table*}[ht!]
\caption{Rank among Subjects under W Stage}
\vspace{0.1in}
\begin{center}
\begin{small}
\begin{sc}
\begin{tabular}{ccccccccccc}
    \toprule
    microstate&\multicolumn{10}{c}{rank among $10$ groups of subjects}\\
    \midrule
    \vdots&\multicolumn{10}{c}{\vdots}\\
    $160$&$24$&$29$&$23$&$23$&$30$&$25$&$26$&$25$&$20$&$27$\\
    $161$&$14$&$11$&$16$&$17$&$14$&$15$&$15$&$14$&$19$&$14$\\
    $162$&$917$&$827$&$892$&$872$&$814$&$775$&$839$&$813$&$807$&$858$\\
    \vdots&\multicolumn{10}{c}{\vdots}\\
    $384$&$229$&$166$&$236$&$165$&$188$&$218$&$150$&$220$&$258$&$165$\\
    $385$&$3$&$2$&$3$&$2$&$4$&$2$&$3$&$2$&$3$&$2$\\
    $386$&$338$&$721$&$589$&$677$&$543$&$640$&$654$&$531$&$402$&$634$\\
    \vdots&\multicolumn{10}{c}{\vdots}\\
    $418$&$701$&$519$&$470$&$667$&$672$&$659$&$597$&$511$&$409$&$587$\\
    $419$&$1$&$1$&$1$&$1$&$1$&$1$&$1$&$1$&$1$&$1$\\
    $420$&$537$&$396$&$645$&$619$&$592$&$608$&$670$&$609$&$607$&$654$\\
    $421$&$2$&$3$&$2$&$3$&$2$&$3$&$2$&$3$&$2$&$3$\\
    $422$&$573$&$434$&$515$&$458$&$496$&$508$&$525$&$494$&$349$&$548$\\
    \vdots&\multicolumn{10}{c}{\vdots}\\
    \bottomrule
\end{tabular}
\end{sc}
\end{small}
\end{center}
\end{table*}

\begin{table*}[ht!]
\caption{Rank among Subjects under R Stage}
\vspace{0.1in}
\begin{center}
\begin{small}
\begin{sc}
\begin{tabular}{ccccccccccc}
    \toprule
    microstate&\multicolumn{10}{c}{rank among $10$ groups of subjects}\\
    \midrule
    \vdots&\multicolumn{10}{c}{\vdots}\\
    $160$&$27$&$23$&$24$&$22$&$26$&$23$&$25$&$25$&$20$&$24$\\
    $161$&$15$&$15$&$18$&$21$&$14$&$14$&$21$&$16$&$22$&$21$\\
    $162$&$831$&$810$&$840$&$699$&$672$&$762$&$772$&$688$&$882$&$720$\\
    \vdots&\multicolumn{10}{c}{\vdots}\\
    $384$&$158$&$156$&$202$&$202$&$168$&$180$&$174$&$192$&$220$&$200$\\
    $385$&$5$&$4$&$7$&$3$&$4$&$3$&$8$&$3$&$3$&$3$\\
    $386$&$544$&$626$&$712$&$818$&$818$&$817$&$773$&$736$&$714$&$634$\\
    \vdots&\multicolumn{10}{c}{\vdots}\\
    $418$&$466$&$639$&$569$&$527$&$634$&$592$&$557$&$594$&$528$&$616$\\
    $419$&$1$&$1$&$2$&$1$&$1$&$1$&$1$&$1$&$1$&$1$\\
    $420$&$589$&$671$&$731$&$581$&$631$&$711$&$753$&$644$&$811$&$722$\\
    $421$&$3$&$2$&$1$&$2$&$2$&$2$&$2$&$2$&$2$&$2$\\
    $422$&$420$&$386$&$368$&$281$&$460$&$332$&$381$&$352$&$325$&$346$\\
    \vdots&\multicolumn{10}{c}{\vdots}\\
    \bottomrule
\end{tabular}
\end{sc}
\end{small}
\end{center}
\end{table*}

From the above tables we see that the $4$ microstates $161,385,419$ and $421$ occur frequently in both R and W stages, with roughly the same ranks. This suggests that the brain undergoes similar activity patterns during these stages.

\begin{figure}[ht]
    \centering
    \includegraphics[width=\linewidth]{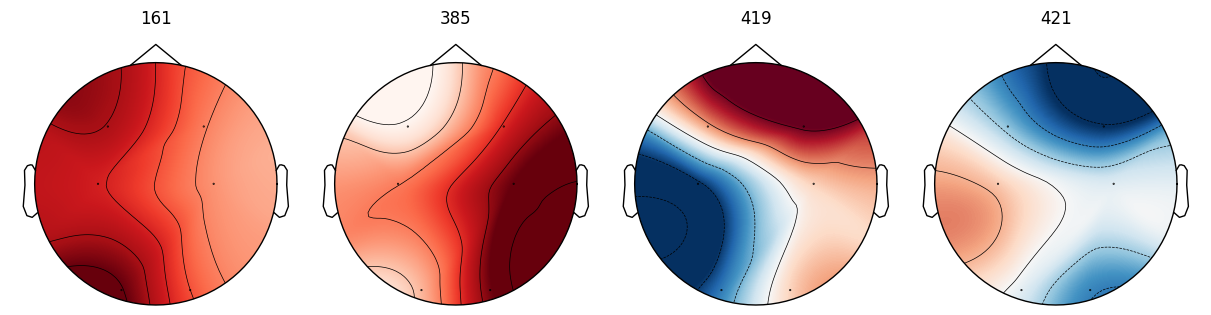}
    \caption{Visualizing Microstates $161,385,419$ and $421$}
    \label{fig:microstate_W_R}
\end{figure}

Further examination of these microstates shows that these microstates have low potential which is below $10\mu$V. This is consistent with the fact that during W stage, brain signals are dominated by $\alpha$ waves which have a low potential. Also, this result indicates certain similarities between W stage and R stage since the brain undergoes similar activity.
\subsection{Comparison Between the Wake Stage and the Non-Rapid Eye Movement III Stage}
NREM$3$ stage denotes deep sleep. In this case, the brain activity differs from that in the wake stage.

\begin{table*}[ht!]
\caption{Rank among Subjects under N$3$ Stage}
\vspace{0.1in}
\begin{center}
\begin{small}
\begin{sc}
\begin{tabular}{ccccccccccc}
    \toprule
    microstate&\multicolumn{10}{c}{rank among $10$ groups of subjects}\\
    \midrule
    \vdots&\multicolumn{10}{c}{\vdots}\\
    $377$&$625$&$396$&$672$&$703$&$134$&$737$&$748$&$577$&$358$&$771$\\
    $378$&$1$&$2$&$3$&$1$&$2$&$2$&$1$&$2$&$2$&$1$\\
    $379$&$800$&$888$&$799$&$845$&$825$&$832$&$830$&$824$&$817$&$791$\\
    \vdots&\multicolumn{10}{c}{\vdots}\\
    $451$&$753$&$811$&$707$&$779$&$627$&$755$&$822$&$871$&$614$&$782$\\
    $452$&$18$&$7$&$10$&$10$&$10$&$15$&$8$&$5$&$13$&$8$\\
    $453$&$939$&$979$&$915$&$881$&$902$&$932$&$709$&$910$&$965$&$899$\\
    \vdots&\multicolumn{10}{c}{\vdots}\\
    $650$&$28$&$148$&$89$&$244$&$117$&$68$&$109$&$170$&$135$&$120$\\
    $651$&$6$&$4$&$13$&$2$&$4$&$6$&$3$&$4$&$3$&$2$\\
    $652$&$913$&$908$&$793$&$786$&$929$&$814$&$850$&$755$&$872$&$784$\\
    \vdots&\multicolumn{10}{c}{\vdots}\\
    \bottomrule
\end{tabular}
\end{sc}
\end{small}
\end{center}
\end{table*}

From the microstates distribution we see that the dominant microstates are different from that of W stage. To further back this observation, we record the rank of microstates $378,452,385$ across W and N$3$ stage.

\begin{table*}[ht!]
\caption{Rank among Subjects under N$3$ Stage}
\vspace{0.1in}
\begin{center}
\begin{small}
\begin{sc}
\begin{tabular}{cccccccccccc}
    \toprule
    \multicolumn{2}{c}{microstate}&\multicolumn{10}{c}{rank among $10$ groups of subjects}\\
    \midrule
    \multirow{2}{*}{$378$}&W&$132$&$120$&$214$&$146$&$169$&$158$&$109$&$165$&$172$&$163$ \\
    &N$3$&$0$&$1$&$2$&$0$&$1$&$1$&$0$&$1$&$1$&$0$ \\
    \midrule
    \multirow{2}{*}{$452$}&W&$95$&$85$&$166$&$98$&$112$&$115$&$112$&$100$&$143$&$132$ \\
    &N$3$&$17$&$6$&$9$&$9$&$9$&$14$&$7$&$4$&$12$&$7$ \\
    \midrule
    \multirow{2}{*}{$421$}&W&$3$&$2$&$1$&$2$&$2$&$2$&$2$&$2$&$2$&$2$ \\
    &N$3$&$489$&$124$&$26$&$52$&$278$&$161$&$71$&$54$&$34$&$78$ \\
    \bottomrule
\end{tabular}
\end{sc}
\end{small}
\end{center}
\end{table*}

Results demonstrate that the microstates that frequently occur during W stage typically occur rarely in N$3$ stage. This again shows that the brain activity differs considerably between these $2$ stages.

We further visualize the microstates:
\begin{figure}[ht]
    \centering
    \includegraphics[width=\linewidth]{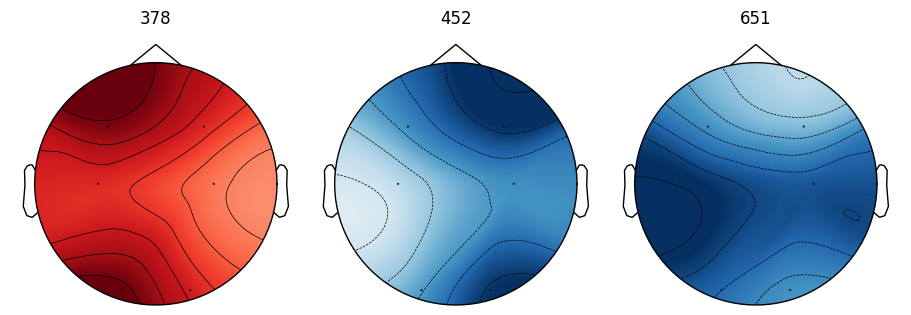}
    \caption{Visualizing Microstates $378,452$ and $651$}
    \label{fig:microstate_N3}
\end{figure}

and we see that these microstates correspond to a state with a relatively high potential, typically $>10\mu$V. This is consistent with the fact that during N$3$ stage, brain activity is dominated by $\delta$ waves which has a high amplitude \cite{ZHANG2024651}. Nonetheless, EEG signals are oscillating and will not always remain at a high voltage, hence in certain cases low-potential states like the microstates dominating in W stage will also occur with a relatively high frequency.

\begin{table}[htbp!]
\centering
\begin{minipage}{0.15\textwidth}
    \centering
    \begin{tabular}{c|c}
        \hline
        microstate & rank \\
        \hline
        $419$ & $1$ \\
        $333$ & $2$ \\
        $421$ & $3$ \\
        \vdots & \vdots \\
        $385$ & $5$ \\
        \vdots & \vdots \\
        $161$&$15$\\
        \vdots&\vdots\\
        \vdots&\vdots\\
        \hline
    \end{tabular}
    \vspace{0.2cm}

    subjects $1\sim30$ R stage
\end{minipage}
\hfill
\begin{minipage}{0.15\textwidth}
    \centering
    \begin{tabular}{c|c}
        \hline
        microstate & rank \\
        \hline
        $419$ & $1$ \\
        $421$&$2$\\
        $333$ & $3$ \\
        $385$ & $4$ \\
        \vdots & \vdots \\
        \vdots & \vdots \\
        $161$ & $11$ \\
        \vdots & \vdots \\
        \vdots&\vdots\\
        \hline
    \end{tabular}
    \vspace{0.2cm}

    subjects $31\sim60$ R stage
\end{minipage}
\hfill
\begin{minipage}{0.15\textwidth}
    \centering
    \begin{tabular}{c|c}
        \hline
        microstate & rank \\
        \hline
        $421$ & $1$ \\
        $419$ & $2$ \\
        $333$ & $3$ \\
        \vdots & \vdots \\
        $385$ & $7$ \\
        \vdots & \vdots \\
    $161$&$18$\\
        \vdots & \vdots \\
        \vdots & \vdots \\
        \hline
    \end{tabular}
    \vspace{0.2cm}

    subjects $61\sim90$ R stage
\end{minipage}
\caption{Rank among Subjects under N$3$ Stage}
\end{table}
\begin{table}[htbp!]
\centering
\begin{minipage}{0.15\textwidth}
    \centering
    \begin{tabular}{c|c}
        \hline
        microstate & rank \\
        \hline
        $419$ & $1$ \\
        $421$ & $2$ \\
        $385$ & $3$ \\
        \vdots & \vdots \\
        $333$ & $7$ \\
        \vdots & \vdots \\
        $161$&$14$\\
        \vdots & \vdots \\
        \vdots & \vdots \\
        \hline
    \end{tabular}
    \vspace{0.2cm}

    subjects $1\sim30$ W stage
\end{minipage}
\hfill
\begin{minipage}{0.15\textwidth}
    \centering
    \begin{tabular}{c|c}
        \hline
        microstate & rank \\
        \hline
        $419$ & $1$ \\
        $385$ & $2$ \\
        $421$ & $3$ \\
        \vdots & \vdots \\
        $333$ & $6$ \\
        \vdots & \vdots \\
        $161$&$11$\\
        \vdots & \vdots \\
        \vdots & \vdots \\
        \hline
    \end{tabular}
    \vspace{0.2cm}

    subjects $31\sim60$ W stage
\end{minipage}
\hfill
\begin{minipage}{0.15\textwidth}
    \centering
    \begin{tabular}{c|c}
        \hline
        microstate & rank \\
        \hline
        $419$ & $1$ \\
        $421$ & $2$ \\
        $385$ & $3$ \\
        \vdots&\vdots\\
        $333$ & $7$ \\
        \vdots & \vdots \\
        $161$&$16$\\
        \vdots & \vdots \\
        \vdots & \vdots \\
        \hline
    \end{tabular}
    \vspace{0.2cm}

    subjects $61\sim90$ W stage
\end{minipage}
\caption{Rank among Subjects under N$3$ Stage}
\end{table}
\begin{table}[htbp!]
\centering
\begin{minipage}{0.15\textwidth}
    \centering
    \begin{tabular}{c|c}
        \hline
        microstate & rank \\
        \hline
        $378$ & $1$ \\
        \vdots&\vdots\\
        $651$ & $6$ \\
        \vdots & \vdots \\
        $487$ & $14$ \\
        \vdots & \vdots \\
        $452$ & $18$ \\
        \vdots & \vdots \\
        \vdots & \vdots \\
        \hline
    \end{tabular}
    \vspace{0.2cm}

    subjects $1\sim30$ N$3$ stage
\end{minipage}
\hfill
\begin{minipage}{0.15\textwidth}
    \centering
    \begin{tabular}{c|c}
        \hline
        microstate & rank \\
        \hline
        \vdots & \vdots \\
        $378$ & $2$ \\
        \vdots&\vdots\\
        $651$ & $4$ \\
        \vdots & \vdots \\
        $452$ & $7$ \\
        \vdots & \vdots \\
        $487$ & $11$ \\
        \vdots & \vdots \\
        \hline
    \end{tabular}
    \vspace{0.2cm}

    subjects $31\sim60$ N$3$ stage
\end{minipage}
\hfill
\begin{minipage}{0.15\textwidth}
    \centering
    \begin{tabular}{c|c}
        \hline
        microstate & rank \\
        \hline
        \vdots&\vdots\\
        $487$ & $2$ \\
        $378$ & $3$ \\
        \vdots & \vdots \\
        $452$ & $10$ \\
        \vdots&\vdots\\
        $651$ & $13$ \\
        \vdots & \vdots \\
        \vdots & \vdots \\
        \hline
    \end{tabular}
    \vspace{0.2cm}

    subjects $61\sim90$ N$3$ stage
\end{minipage}
\caption{Rank among Subjects under N$3$ Stage}
\end{table}

\subsection{Comparing other Sleep Stages}
For sleep stage N$1$ and N$2$, the dominant microstates are also $419,385,421,615$ and other states found in W stage. This suggests that these microstates capture a class of weak EEG signals that the brain usually switches between. Also, the brain activity in stages N$1$ and N$2$ shares certain aspects with that in W stage.

We also found that when transforming from stage W through stage N$1$, stage N$2$ and finally to stage N$3$, the frequency of microstate $489$ is increasing. The visualization of $489$ is as follows:
\begin{figure}[ht]
    \centering
    \includegraphics[width=0.3\linewidth]{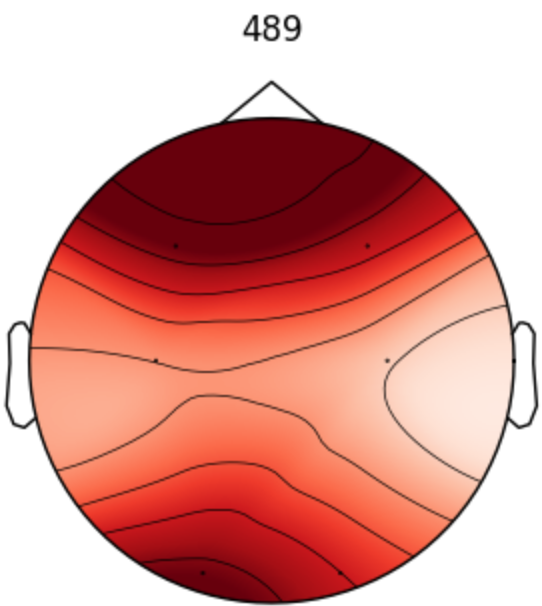}
    \caption{Visualizing Microstates $489$}
    \label{fig:microstate_489}
    \Description{489}
\end{figure}

which also has relatively high potential with all leads between $2\sim14\mu$V. This again shows that from W through N$1$, N$2$ to N$3$, high amplitude brain activity becomes more and more common.

\end{document}